\documentclass{article}

     \usepackage[final]{neurips_2024}

\usepackage[utf8]{inputenc} 
\usepackage[T1]{fontenc}    
\usepackage{url}            
\usepackage{booktabs}       
\usepackage{amsfonts}       
\usepackage{nicefrac}       
\usepackage{microtype}      
\usepackage{xcolor}         
\usepackage{bm} 
\usepackage{tabularx}
\newcolumntype{Y}{>{\centering\arraybackslash}X}
\usepackage{multirow}
\newcommand{\dsname}[1]{D-Rep}
\newcommand{\msname}[1]{PDF-Embedding}
\newcommand{\tsname}[1]{ICDiff}
\usepackage{graphicx}
\usepackage{multicol}
\usepackage{comment}
\usepackage{xcolor}
\usepackage{mdframed}
\usepackage{caption}
\usepackage{amsmath}
\usepackage{float}
\usepackage{colortbl}
\usepackage[linkcolor=red, citecolor=green, urlcolor=blue, colorlinks=true]{hyperref} 
\title{Image Copy Detection for Diffusion Models}

\author{%
  Wenhao Wang \\
  ReLER Lab, University of Technology Sydney\\
  \texttt{wangwenhao0716@gmail.com} \\
  \And
  Yifan Sun\thanks{Corresponding Author.} \\
  Baidu Inc. \\
  \texttt{sunyifan01@baidu.com} \\
  \And  
  \hspace{1.5cm}Zhentao Tan \\
  \hspace{1.5cm}AAIS, Peking University \\
  \hspace{1.5cm}\texttt{tanzhentao@stu.pku.edu.cn} \\
  \And
  \hspace{0.9cm}Yi Yang \\
  \hspace{0.9cm}ReLER Lab, Zhejiang University\\
  \hspace{0.9cm}\texttt{yangyics@zju.edu.cn} \\
}
\author{
Wenhao Wang\textsuperscript{\rm 1}, 
Yifan Sun\textsuperscript{\rm 2}\thanks{Corresponding Author.} ,
Zhentao Tan\textsuperscript{\rm 2},
Yi Yang\textsuperscript{\rm 3}\\
\textsuperscript{\rm 1}University of Technology Sydney
\textsuperscript{\rm 2}Baidu Inc.
\textsuperscript{\rm 3}Zhejiang University \\
}

\begin{document}
\maketitle

\begin{center}
    \centering
    \vspace{-2.0em}
    \includegraphics[width=14cm]{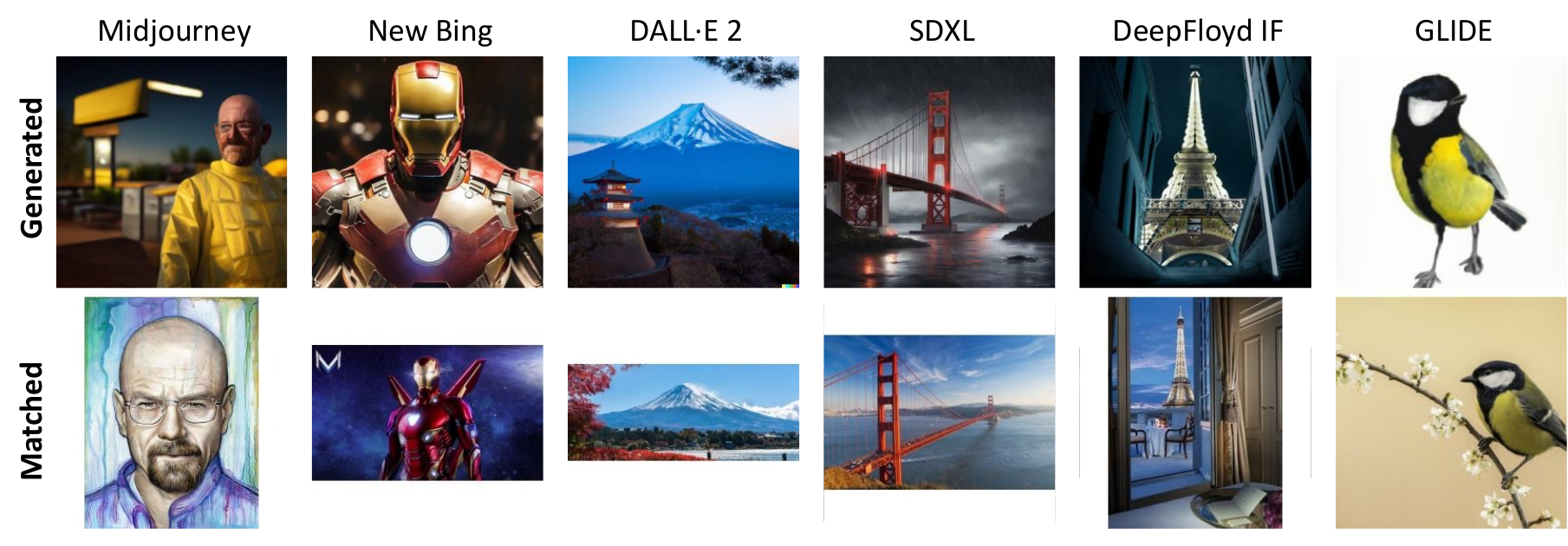}
    \captionof{figure}{Some generated images (top) from diffusion models replicates the contents of existing images (bottom). The existing (matched) images are from LAION-Aesthetics \cite{schuhmann2022laion}. The diffusion models include both commercial and open-source ones.}
    \label{Fig: famous}
\vspace{3mm}
\end{center}

\begin{abstract}
Images produced by diffusion models are increasingly popular in digital artwork and visual marketing. 
However, such generated images might replicate content from existing ones and pose the challenge of content originality. 
Existing Image Copy Detection (ICD)  models, though accurate in detecting hand-crafted replicas, overlook the challenge from diffusion models. This motivates us to introduce \tsname~, the first ICD specialized for diffusion models. To this end, we construct a Diffusion-Replication (\dsname~) dataset and correspondingly propose a novel deep embedding method. \dsname~ uses a state-of-the-art diffusion model (Stable Diffusion V1.5) to 
generate $40,000$ image-replica pairs, which are manually annotated into 6 replication levels ranging from $0$ (no replication) to $5$ (total replication). Our method, \msname~, transforms the replication level of each image-replica pair into a probability density function (PDF) as the supervision signal. The intuition is that the probability of neighboring replication levels should be continuous and smooth. Experimental results show that \msname~ surpasses protocol-driven methods and non-PDF choices on the \dsname~ test set. 
Moreover, by utilizing \msname~, we find that the replication ratios of well-known diffusion models against 
an open-source gallery range from $10\%$ to $20\%$. The project is publicly available at \url{https://icdiff.github.io/}.
\end{abstract}

\section{Introduction}
\label{sec:intro}
\begin{figure}
    \includegraphics[width=10cm]{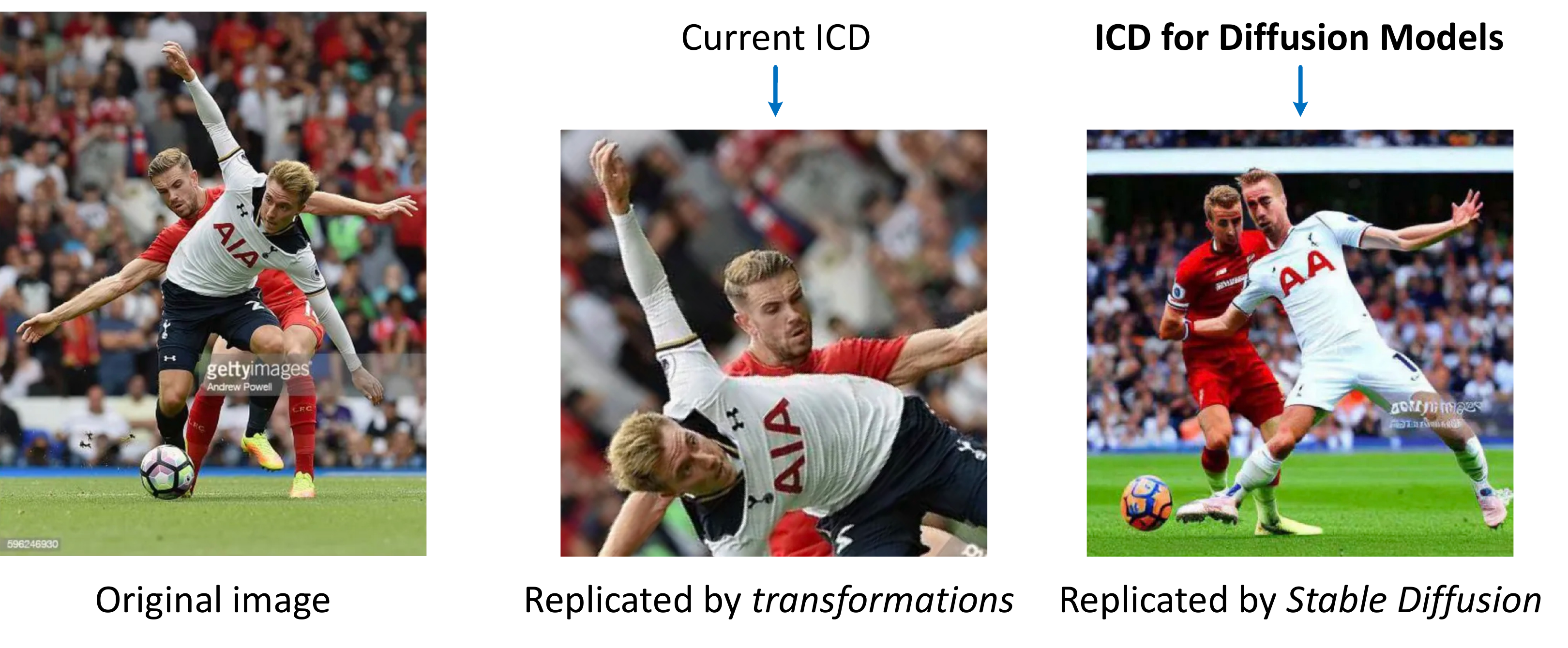}
    \centering
    \captionof{figure}{The comparison between current ICD with the \tsname~. The current ICD focuses on detecting edited copies generated by transformations like horizontal flips, random rotations, and random crops. In contrast, the \tsname~ aims to detect replication generated by diffusion models, such as Stable Diffusion \cite{rombach2021highresolution}. (Source of the original image: \href{https://www.theverge.com/2023/2/6/23587393/ai-art-copyright-lawsuit-getty-images-stable-diffusion}{Lawsuit from Getty Images.})} 
    \label{Fig: trad_diff}
\end{figure}

Diffusion models have gained popularity due to their ability to generate high-quality images. 
A phenomenon accompanying this trend is that these generated images might replicate content from existing ones. In Fig. \ref{Fig: famous}, we choose 
six well-known diffusion models \cite{midjourney2022,new_bing,ramesh2022hierarchical, podell2023sdxl,deep_floyd_IF,nichol2022glide} to illustrate this replication phenomenon. The content replication is acceptable for some (fair) use while interest holders may regard others as copyright infringement \cite{lee2024talkin,somepalli2023diffusion,wen2024detecting}. This paper leaves this dispute alone, and focuses a scientific problem: \textit{How to identify the content replication brought by diffusion models?}



Image Copy Detection (ICD) provides a general solution to the above demand: it identifies whether an image is copied from a reference gallery after being tampered with. However, the current ICD methods are trained using hand-crafted image transformations (\emph{e.g.}, horizontal flips, random rotations, and random crops) and overlook the challenge from diffusion models. Empirically, we find existing ICD methods can be easily confused by diffusion-generated replicas (as detailed in Table \ref{Table: plausible}). We infer it is because the tamper patterns underlying diffusion-generated replicas (Fig.~\ref{Fig: trad_diff} right) are different from hand-crafted ones (Fig.~\ref{Fig: trad_diff} middle), yielding a considerable pattern gap. 

In this paper, we introduce \tsname~, the first ICD specialized for diffusion-generated replicas. Our efforts mainly involve building a new ICD dataset and proposing a novel deep embedding method. 

$\bullet$ \textbf{A Diffusion Replication (\dsname~) dataset.} \dsname~ consists of $40,000$ image-replica pairs, in which each replica is generated by a diffusion model. Specifically, the images are from LAION-Aesthetic V2 \cite{schuhmann2022laion}, while their replicas are generated by Stable Diffusion V1.5 \cite{rombach2022high}. To make the replica generation more efficient, we search out the text prompts (from DiffusionDB \cite{wang-etal-2023-diffusiondb})that are similar to the titles of LAION-Aesthetic V2 images, input these text prompts into Stable Diffusion V1.5, and generate many redundant candidate replicas. Given these candidate replicas, we employ human annotators to label the replication level of each generated image against a corresponding LAION-Aesthetic image. 
The annotation results in $40,000$ image-replica pairs with $6$ replication levels ranging from $0$ (no replication) to $5$ (total replication). 
We divide \dsname~ into a training set with $90\%$ ($36,000$) pairs and a test set with the remaining $10\%$ ($4,000$) pairs. 

$\bullet$ \textbf{A novel method named PDF-Embedding.} The ICD methods rely on deep embedding learning at their core. In the deep embedding space, the replica should be close to its original image and far away from other images. Compared with popular deep embedding methods, our \msname~ learns a Probability-Density-Function between two images, instead of a similarity score. More concretely, \msname~ transforms the replication level of each image-replica pair into a PDF as the supervision signal. The intuition is that the probability of neighboring replication levels should be continuous and smooth. For instance, if an image-replica pair is annotated as level-$3$ replication, the probabilities for level-$2$ and level-$4$ replications should also not be significantly low. 

\msname~ predicts the probability scores on all the replication levels simultaneously in two steps:
1) extracting $6$ feature vectors in parallel from both the real image and its replica, respectively and 2) calculating $6$ inner products (between two images) to indicate the probability score at $6$ corresponding replication levels. The largest-scored entry indicates the predicted replication level. 
Experimentally, we prove the effectiveness of our method by comparing it with popular deep embedding models and protocol-driven methods trained on our \dsname~.  Moreover, we evaluate the replication of six famous diffusion models and provide a comprehensive analysis.\par 

In conclusion, our key contributions are as follows:
\begin{enumerate}
 \item We propose a timely and important ICD task, \textit{i.e}, Image Copy Detection for Diffusion Models (\tsname~), designed specifically to identify the replication caused by diffusion models. 
 \item We build the first \tsname~ dataset and introduce \msname~ as a baseline method. \msname~ transforms replication levels into probability density functions (PDFs) and learns a set of representative vectors for each image.
 \item Extensive experimental results demonstrate the efficiency of our proposed method. Moreover, we discover that between $10\%$ to $20\%$ of images generated by six well-known diffusion models replicate contents of a large-scale image gallery.
\end{enumerate}
\par 

\section{Related Works}
\label{sec: related}

\subsection{Existing Image Copy Detection Methods}
Current ICD methods try to detect replications by learning the invariance of image transformations. For example, ASL \cite{wang2023benchmark} considers the relationship between image transformations and hard negative samples. AnyPattern \cite{wang2024AnyPattern} and PE-ICD \cite{wang2024peicd} build benchmarks and propose solutions that focus on novel patterns in real-world scenarios. SSCD \cite{pizzi2022self} reveals that self-supervised contrastive training inherently relies on image transformations, and thus adapts InfoNCE \cite{oord2018representation} by introducing a differential entropy regularization. BoT \cite{wang2021bag} incorporates approximately ten types of image transformations, combined with various tricks, to train an ICD model. By increasing the intensity of transformations gradually, CNNCL \cite{yokoo2021contrastive} successfully detects hard positives using a simple contrastive loss and memory bank. EfNet \cite{papadakis2021producing} ensembles models trained with different image transformations to boost the performance.
In this paper, we discover that capturing the invariance of image transformations is ineffective for detecting copies generated by diffusion models. Consequently, we manually label a new dataset and train a specialized ICD model.

\subsection{Replication in Diffusion Models}
Past research has explored the replication problems associated with diffusion models. The study by \cite{somepalli2023diffusion} questions if diffusion models generate unique artworks or simply mirror the content from their training data. Research teams from Google, as highlighted in \cite{carlini2023extracting}, note that diffusion models reveal their training data during content generation. Other studies prevent generating replications from the perspectives of both training diffusion models \cite{zhang2023forget,somepalli2023understanding,vyas2023provable,anonymous2023copyright,zhao2023recipe} and copyright holders \cite{cui2023ft,rhodes2023my,cui2023diffusionshield}. Some experts, such as those in \cite{somepalli2023understanding}, find that the replication of data within training sets might be a significant factor leading to copying behaviors in diffusion models. To address this, \cite{webster2023duplication} proposes an algorithmic chain to de-duplicate the training sources like LAION-2B \cite{schuhmann2022laion}. 
In contrast to these efforts, our \tsname~ offers a unique perspective. Specifically, unlike those that directly using existing image descriptors (such as those from CLIP \cite{radford2021learning} and SSCD \cite{pizzi2022self}), we manually-label a dataset and develop a specialized ICD algorithm. By implementing our method, the analytical tools and preventative strategies proposed in existing studies may achieve greater efficacy.

\section{Benchmark}
\begin{figure*}[t]
    \includegraphics[width=14cm]{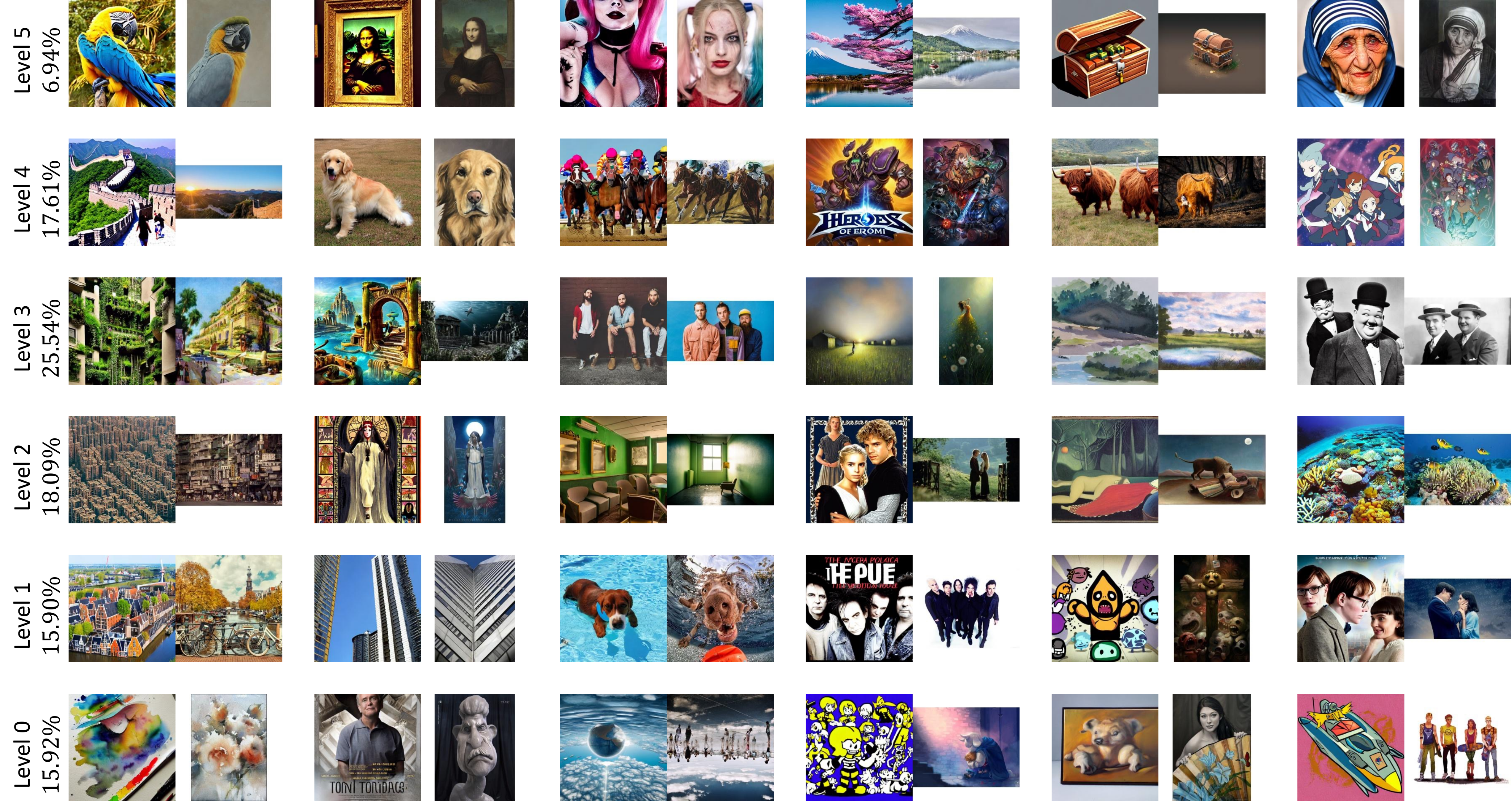}
    \captionof{figure}{The demonstration of the manual-labeled \dsname~ dataset. The percentages on the left show the proportion of images with a particular level.} 
\vspace{-1mm}
    \label{Fig: label}
\end{figure*}

This section introduces the proposed ICD for diffusion models (\tsname~), including our dataset (\dsname~) and the corresponding evaluation protocols.
\subsection{\dsname~ Dataset}
Current ICD \cite{douze2009evaluation,douze20212021,papakipos2022results,wang2023benchmark,wang2024peicd,wang2024AnyPattern} primarily focuses on the replica challenges brought by hand-crafted transformations. In contrast, our \tsname~ aims to address the replication issues caused by diffusion models \cite{midjourney2022,new_bing,ramesh2022hierarchical, podell2023sdxl,deep_floyd_IF,nichol2022glide}. 
To facilitate \tsname~ research, we construct \dsname~ dataset, which is characterized for diffusion-based replication (See Fig. \ref{Fig: label} and the Appendix (Section \ref{supple: more_example}) for the examples of diffusion-based replica).
The construction process involves generating candidate pairs followed by manual labeling. \par 



\textbf{Generating candidate pairs.} It consists of (1) selecting the top $40,000$ most similar prompts and titles: this selection provides an abundant image-replica pair source.  In detail, we use the Sentence Transformer \cite{reimers2019sentence} to encode the $1.8$ million real-user generated prompts from DiffusionDB \cite{wang-etal-2023-diffusiondb} and the $12$ million image titles from LAION-Aesthetics V2 6+ \cite{schuhmann2022laion}, and then utilize the computed cosine similarities to compare; (2) obtaining the candidate pairs: the generated images are produced using the prompts with Stable Diffusion V1.5 \cite{rombach2022high}, and the real images are fetched based on the titles.


\textbf{Manual labeling.} We generally follow the definition of replication in \cite{somepalli2023diffusion} and further define six levels of replication (0 to 5). A higher level indicates a greater degree that the generated image replicates the real image. Due to the complex nature of diffusion-generated images, we use multiple levels instead of the binary levels used in \cite{somepalli2023diffusion}, which employed manual-synthetic datasets as shown in their Fig. 2. 
We then train ten professional labelers to assign these levels to the $40,000$ candidate pairs:
Initially, we assign $4,000$ image pairs to each labeler. If labelers are confident in their judgment of an image pair, they will directly assign a label. Otherwise, they will place the image pair in an undecided pool. On average, each labeler has about $600$ undecided pairs. Finally, for each undecided pair, we vote to reach a final decision. For example, if the votes for an undecided pair are $2$, $2$, $2$, $3$, $3$, $3$, $3$, $3$, $4$, $4$, the final label assigned is $3$.
Given the complexity of this labeling task, it took both the labelers and our team one month to finish the process. To maintain authenticity, we did not pre-determine the proportion of each score. The resulting proportions are on the left side of Fig. \ref{Fig: label}. 


\subsection{Evaluation Protocols}
To evaluate ICD models on the \dsname~ dataset, we divide the dataset into a 90/10 training/test split and design two evaluation protocols: Pearson Correlation Coefficient (PCC) and Relative Deviation (RD). \par 

\textbf{Pearson Correlation Coefficient (PCC).} The PCC is a measure used to quantify the linear relationship between two sequences. When PCC is near $1$ or $-1$, it indicates a strong positive or negative relationship. If PCC is near $0$, there's little to no correlation between the sequences. Herein, we consider two sequences, the predicted replication level $\boldsymbol{s}^p=\left(s_1^p, s_2^p, \ldots, s_n^p\right)$ and the ground-truth $\boldsymbol{s}^l=\left(s_1^l, s_2^l, \ldots, s_n^l\right)$ ($n$ is the number of test pairs). The PCC for \tsname~ is defined as:
\begin{equation}
\text{PCC}=\frac{\sum\limits_{i=1}^{n} \left( s^{p}_{i}-\bar{\bm{s}^{p} } \right)  \left( s^{l}_{i}-\bar{\bm{s}^{l} } \right)  }{\sqrt{\sum\limits_{i=1}^{n} \left( s^{p}_{i}-\bar{\bm{s}^{p} } \right)^{2}  } \times \sqrt{\sum\limits_{i=1}^{n} \left( s^{l}_{i}-\bar{\bm{s}^{l} } \right)^{2}  } },
\end{equation}
where $\bar{\bm{s}^{p} }$ and $\bar{\bm{s}^{l} }$ are the mean values of $\bm{s}^{p}$ and $\bm{s}^{l}$, respectively. \par 
A limitation of the PCC is its insensitivity to global shifts. If all the predictions differ from their corresponding ground truth with the same shift, the PCC does not reflect such a shift and remains large. To overcome this limitation, we propose a new metric called the Relative Deviation (RD). \par 


\textbf{Relative Deviation (RD).} We use RD to quantify the normalized deviation between the predicted and the labeled levels. By normalizing against the maximum possible deviation, RD provides a measure of how close the predictions are to the labeled levels on a scale of $0$ to $1$. The RD is calculated by
\begin{equation}
\text{RD}=\frac{1}{n} \sum^{n}_{i=1} \left( \frac{\left| s^{p}_{i}-s^{l}_{i}\right|  }{\max \left( N-s^{l}_{i},s^{l}_{i}\right)  } \right)  ,
\end{equation}
where $N$ is the highest replication level in our \dsname~. \par 

\textbf{The Preference for RD over Absolute One.} Here we show the preference for employing RD over the absolute one through two illustrative examples. We denote the relative and absolute deviation of the $i$th test pair as:
$
S_{i}=\frac{\left| s^{p}_{i}-s^{l}_{i}\right|  }{\max \left( N-s^{l}_{i},s^{l}_{i}\right)  } ,
$
and 
$
T_{i}=\frac{\left| s^{p}_{i}-s^{l}_{i}\right|  }{N} .
$\par 
(1) For a sample with $s^{l}_{i}=3$, if $s^{p}_{i}=3$, both $S_i$ and $T_i$ equal $0$; however, if $s^{p}_{i}=0$ (representing the worst prediction), $S_i=1$ and $T_i=0.6$. Here, $S_i$ adjusts the worst prediction to a value of $1$. \par 
(2) In the first scenario, where $s^{l}_{i}=3$ and $s^{p}_{i}=0, S_i=1$ and $T_i=0.6$. In the second scenario, where $s^{l}_{i}=5$ and $s^{p}_{i}=2, S_i=0.6$ and $T_i=0.6$. For both cases, $T_i$ remains the same at $0.6$, whereas $S_i$ values differ. Nevertheless, the two scenarios are distinct: in the first, the prediction cannot deteriorate further; in the second, it can. The value of $S_i$ accurately captures this distinction, whereas $T_i$ does not. \par

\section{Method}
\begin{figure*}[t]
    \includegraphics[width=14cm]{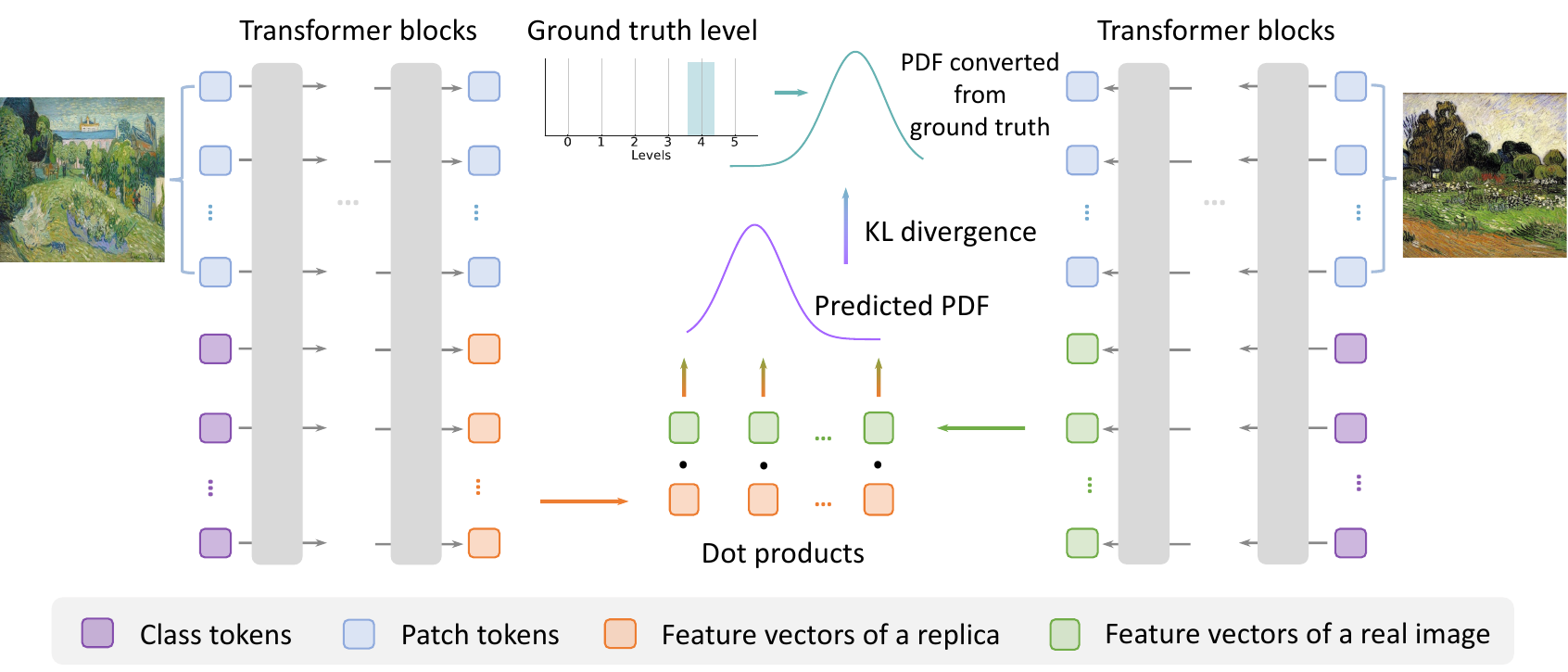}
    \captionof{figure}{The demonstration of the proposed \msname~. Initially, \msname~ converts manually-labeled replication levels into probability density functions (PDFs). To learn from these PDFs, we use a set of vectors as the representation of an image.} 
\vspace{-1mm}
    \label{Fig: DiffICD}
\end{figure*}

This section introduces our proposed \msname~ for \tsname~. \msname~ converts each replication level into a probability density function (PDF). To facilitate learning from these PDFs, we expand the original representative vector into a set of vectors. The demonstration of \msname~ is displayed in Fig. \ref{Fig: DiffICD}.

\subsection{Converting Level to Probability Density}\label{Sec: req}
Given an replication level $s^l \in \boldsymbol{s}^l$, we first normalize it into $p^l=s^l/ max(\boldsymbol{s}^l)$. Then we transfer $p^l$ into a PDF denoted as $g(x)$
\footnote{Although we acknowledge that the random variables in this context are discrete, we still utilize the term ``PDF” to effectively communicate our intuition and present our method under ideal conditions.}, where $x \in[0,1]$ indicates each possible normalized level. The intuition is that the probability distribution of neighboring replication levels should be continuous and smooth. The function $g(x)$ must satisfy the following conditions: (1) $\int^{1}_{0} g(x)dx=1, $ ensuring that $g(x)$ is a valid PDF; (2) $g(x) \geq 0, $ indicating the non-negativity of the PDF; and (3) $ g(x) \leq g(p^l), $ which implies that the density is maximized at the normalized level $p^l$. We use three different implementations for $g(x)$:\par 
\noindent Gaussian: 
\begin{equation}\label{Eq: g}
g(x \mid A, \mu, \sigma)=A \cdot \exp \left(-\frac{(x-\mu)^2}{2 \cdot \sigma^2}\right),
\end{equation}
linear:
\begin{equation}\label{Eq: l}
g(x \mid A, \mu, \beta)=A-\beta \cdot|x-\mu|,
\end{equation}
and exponential:
\begin{equation}\label{Eq: e}
g(x \mid A, \mu, \lambda)=A \cdot \lambda \cdot \exp (-\lambda \cdot|x-\mu|),
\end{equation}
where: $A$ is the amplitude, and $\mu$ is the center; $\sigma$ is the standard deviation of Gaussian function, $\beta$ is the slope of linear function, and $\lambda$ is the spread of exponential function. \par 
The performance achieved with various converted PDFs is illustrated in the experimental section. For additional details, please see the Appendix (Section \ref{supple: pdf}), which includes (1) \textbf{the methodology for calculating distribution values}, (2) \textbf{the visualization of the learned distributions corresponding to different image pairs}, and (3) \textbf{an analysis of the deviation rate from peak values}.

\subsection{Representing an Image as a Set of Vectors}
To facilitate learning from the converted PDFs, we utilize a Vision Transformer (ViT) \cite{dosovitskiy2020image} to represent an image as a set of vectors. Let's denote the patch tokens from a real image as $\mathbf{X}_r^0$ and from a generated image as $\mathbf{X}_g^0$, the ViT model as $f$, and the number of layers in ViT as $L$. The feed-forward process can be expressed as:

\begin{equation}
\begin{aligned}
& {\left[\mathbf{C}_r^L, \mathbf{X}_r^L\right]=f\left(\left[\mathbf{C}^0, \mathbf{X}_r^0\right]\right),} \\
& {\left[\mathbf{C}_g^L, \mathbf{X}_g^L\right]=f\left(\left[\mathbf{C}^0, \mathbf{X}_g^0\right]\right),}
\end{aligned}
\end{equation}
where $\mathbf{C}^0$ is a set of class tokens; $\mathbf{C}_r^L$ is a set of representative vectors for the real images, consisting of vectors $c_{0, r}^L, c_{1, r}^L, \ldots, c_{N, r}^L$, and $\mathbf{C}_g^L$ is a set of representative vectors for the generated images, consisting of vectors $c_{0, g}^L, c_{1, g}^L, \ldots, c_{N, g}^L$; $N$ is the highest replication level.\par 
Therefore, we can use another PDF $h\left( x\right)$ ($x\in [0,1]$) to describe the predicted replication between two images by
\begin{equation}
h(x)=\mathbf{C}_r^L \cdot \mathbf{C}_g^L,
\end{equation}
which expands to:
\begin{equation}\label{Eq: hx}
h(x)=\left[\left\langle\mathbf{c}_{0, r}^L, \mathbf{c}_{0, g}^L\right\rangle,\left\langle\mathbf{c}_{1, r}^L, \mathbf{c}_{1, g}^L\right\rangle, \ldots,\left\langle\mathbf{c}_{N, r}^L, \mathbf{c}_{N, g}^L\right\rangle\right],
\end{equation}
where $\langle\cdot, \cdot\rangle$ denotes the cosine similarity.\par 
For training, recalling the PDF $g(x)$ derived from the level, we define the final loss using the Kullback-Leibler (KL) divergence:
\begin{equation}
\mathcal{L}=D_{K L}(g \| h)=\int_0^1 g(x) \log \left(\frac{g(x)}{h(x)}\right) d x,
\end{equation}
which serves as a measure of the disparity between two probability distributions. Additionally, in the Appendix (Section \ref{supple: learn}), we demonstrate what the network captures during its learning phase. \par 

During testing, the normalized level between two images is denoted by $\hat{p}^{l}$, satisfying $h(x) \leq h(\hat{p}^{l})$. As illustrated in Eqn. \ref{Eq: hx}, $h(x)$ in practice is discrete within the interval $[0,1]$. Consequently, the resulting level is
\begin{equation}
j=\operatorname{argmax} h(x),
\end{equation}
and the normalized level is quantified as $\frac{j}{N}$.

\section{Experiments}
\subsection{Training Details}
We implement our \msname~ using PyTorch \cite{paszke2019pytorch} and distribute its training over 8 A100 GPUs. The ViT-B/16 \cite{dosovitskiy2020image} serves as the backbone and is pre-trained on the ImageNet dataset \cite{deng2009imagenet} using DeiT \cite{touvron2021training}, unless specified otherwise. We resize images to a resolution of $224 \times 224$ before training. A batch size of $512$ is used, and the total training epochs is $25$ with a cosine-decreasing learning rate.

\subsection{Challenge from the \tsname~ Task}

\begin{table}[t]
\caption{The performance of publicly available models and our \msname~ on the \dsname~. For qualitative results, please refer to Section \ref{supple: similarities} in the Appendix.} 
\vspace*{1mm}
\small
  \begin{tabularx}{\hsize}{>{\centering\arraybackslash}p{2cm}|>{\centering\arraybackslash}p{4cm}|Y|Y}
    \hline
    \multirow{1 }{*}{Class}& Method & PCC ($\%$) \( \uparrow \) & RD ($\%$) \( \downarrow \) \\ \hline
    \multirow{3 }{*}{Vision-}& SLIP \cite{mu2022slip} & $31.8$ & $49.7$ \\
    \multirow{3}{*}{language}& BLIP \cite{li2022blip} & $34.8$ & $41.6$ \\
    \multirow{3}{*}{Models}& ZeroVL \cite{cui2022contrastive} & $36.3$ & $36.5$ \\ 
    & CLIP \cite{radford2021learning} & $36.8$ & $35.8$ \\ 
    & GPT-4V \cite{openai2023gpt4} & \textcolor{blue}{$47.3$} & $38.7$ \\ 
    \hline
    \multirow{2 }{*}{Self-}& SimCLR \cite{chen2020simple} & $7.2$ & $49.4$ \\
    \multirow{2 }{*}{supervised}& MAE \cite{he2022masked} & $20.7$ & $67.6$ \\
    \multirow{2 }{*}{Learning}& SimSiam \cite{chen2021exploring} & $33.5$ & $45.4$ \\
    \multirow{2 }{*}{Models}& MoCov3 \cite{he2020momentum} & $35.7$ & $40.3$ \\ 
    & DINOv2 \cite{oquab2023dinov2} & $39.0$ &  \textcolor{blue}{$32.9$}  \\ 
    \hline
    \multirow{3 }{*}{Supervised} & EfficientNet \cite{tan2019efficientnet} & $24.0$ & $59.3$ \\
    \multirow{3 }{*}{Pre-trained} & Swin-B \cite{liu2021swin} & $32.5$ & $38.4$ \\
    \multirow{3 }{*}{Models} & ConvNeXt \cite{liu2022convnet} & $33.8$ & $36.0$ \\ 
    & DeiT-B \cite{touvron2021training} & $35.3$ & $41.7$ \\
    & ResNet-50 \cite{he2016deep} & $37.5$ & $34.5$ \\ 
    \hline
    \multirow{3 }{*}{Current}& ASL \cite{wang2023benchmark} & $5.6$ & $78.1$ \\ 
    \multirow{3 }{*}{ICD}& CNNCL \cite{yokoo2021contrastive} & $19.1$ & $51.7$ \\ 
    \multirow{3 }{*}{Models}& SSCD \cite{pizzi2022self} & $29.1$ & $62.3$ \\
    & EfNet \cite{papadakis2021producing} & $30.5$ & $62.8$ \\
    & BoT \cite{wang2021bag} & $35.6$ & $53.8$ \\
    \hline
  \end{tabularx}
  \label{Table: plausible}
  \vspace*{-3mm}
\end{table}

This section benchmarks popular public models on our \dsname~ test dataset. As Table \ref{Table: plausible} shows, we conduct experiments extensively on vision-language models, self-supervised models, supervised pre-trained models, and current ICD models. We employ these models as feature extractors and calculate the cosine similarity between pairs of image features (except for GPT-4V Turbo \cite{openai2023gpt4}, see Section \ref{supple: gpt} in the Appendix for the implementation of it). For the computation of PCC and RD, we adjust the granularity by scaling the computed cosine similarities by a factor of $N$. In the Appendix (Section \ref{supple: similarities}), we further present the concrete similarities predicted by these models and provide corresponding analysis. We observe that: 
(1) the large multimodal model GPT-4V Turbo \cite{openai2023gpt4} performs best in PCC, while the self-supervised model DINOv2 \cite{oquab2023dinov2} excels in RD. This can be attributed to their pre-training on a large, curated, and diverse dataset. Nevertheless, their performance remains somewhat limited, achieving only $47.3\%$ in PCC and $32.9\%$ in RD. This underscores that even the best publicly available models have yet to effectively address the \tsname~ task.
(2) Current ICD models, like SSCD \cite{pizzi2022self}, which are referenced in analysis papers \cite{somepalli2023diffusion,somepalli2023understanding} discussing the replication issues of diffusion models, indeed show poor performance. For instance, SSCD \cite{pizzi2022self} registers only $29.1\%$ in PCC and $62.3\%$ in RD. Even the more advanced model, BoT \cite{wang2021bag}, only manages $35.6\%$ in PCC and $53.8\%$ in RD. These results underscore the need for a specialized ICD method for diffusion models. Adopting our specialized ICD approach will make their subsequent analysis more accurate and convincing. 
(3) Beyond these models, we also observe that others underperform on the \tsname~ task. This further emphasizes the necessity of training a specialized \tsname~ model. 

\subsection{The Effectiveness of \msname~}

This section demonstrates the effectiveness of our proposed \msname~ by (1) contrasting it against protocol-driven methods and non-PDF choices on the \dsname~ dataset, (2) comparing between different distributions, and (3)  comparing with other models in generalization settings. \par 

\textbf{Comparison against protocol-driven methods.} 
Since we employ PCC and RD as the evaluation protocols, a natural embedding learning would be directly using these protocols as the optimization objective, \emph{i.e.}, enlarging PCC and reducing RD. Moreover, we add another variant of ``reducing RD'', \emph{i.e.}, reducing the absolute deviation $\left|s_i^p-s_i^l\right|$ in a regression manner. The comparisons are summarized in Table \ref{Table: demo}, from which we draw three observations as below: 
(1) Training on \dsname~ with the protocol-driven method achieves good results on their specified protocol but performs bad for the other. While ``Enlarging PCC” attains a commendable PCC, its RD of $40.1\%$ indicates large deviation from the ground truth. ``Reducing RD” or ``Reducing Deviation” shows a relatively good RD ($28.1\%$); however, they exhibit small PCC values that indicate low linear consistency.
(2) Our proposed \msname~ surpasses these protocol-driven methods in both PCC and RD. Compared against ``Enlarging PCC”, our method improves PCC by $1.6\%$ and decreases RD by $16.1\%$. Besides, our method achieves $+16.0\%$ PCC and $-4.1\%$ RD compared against ``Reducing RD” and ``Reducing Deviation”.
(3) The computational overhead introduced by our method is negligible. First, compared to other options, our method only increases the training time by $5.8\%$. Second, our method introduces minimal additional inference time. Third, while our method requires a longer matching time, its magnitude is close to $10^{-9}$, which is negligible when compared to the inference time's magnitude of $10^{-3}$. Further discussions on the matching time in real-world scenarios can be found in Section \ref{Sec: real}.\par 

\begin{table*}[t]
\caption{Our method demonstrates performance superiority over others.} 

\small
  \begin{tabularx}{\hsize}{Y|>{\centering\arraybackslash}p{1.4cm}|>{\centering\arraybackslash}p{1.3cm}|>{\centering\arraybackslash}p{2.1cm}|>{\centering\arraybackslash}p{2.0cm}|>{\centering\arraybackslash}p{2.3cm}}
    \hline
Method&PCC ($\%$) $\uparrow$&RD ($\%$) $\downarrow$&Train ($s/iter$) $\downarrow$&Infer ($s/img$) $\downarrow$&Match ($s/pair$) $\downarrow$\\ \hline
Enlarging PCC&$54.4$&$40.1$&$0.293$& & \\  
Reducing RD &$15.1$&$29.9$&$0.294$&$2.02 \times 10^{\mathbf{-3}}$&$1.02 \times 10^{\mathbf{-9}}$\\  
Regression&$40.3$&$28.1$&$0.292$&&\\  
\hline 
One-hot Label&$37.6$&$43.3$&\multirow{2 }{*}{$0.310$}&\multirow{2 }{*}{$2.07 \times 10^{\mathbf{-3}}$}&\multirow{2 }{*}{$6.97 \times 10^{\mathbf{-9}}$}\\  
Label Smoothing&$35.0$&$36.1$& & &\\  
\hline 
Ours (Gaussian)&$53.7$&\textbf{24.0}& & & \\ 
Ours (Linear)&$54.0$&$24.6$&$0.310$& $2.07 \times 10^{\mathbf{-3}}$&$6.97 \times 10^{\mathbf{-9}}$ \\
Ours (Exp.)&\textbf{56.3}&$25.6$& &&\\ \hline 
  \end{tabularx}
  \label{Table: demo}
  \\
\end{table*}

\begin{figure*}[t]
\centering
    \includegraphics[width=14.5cm]{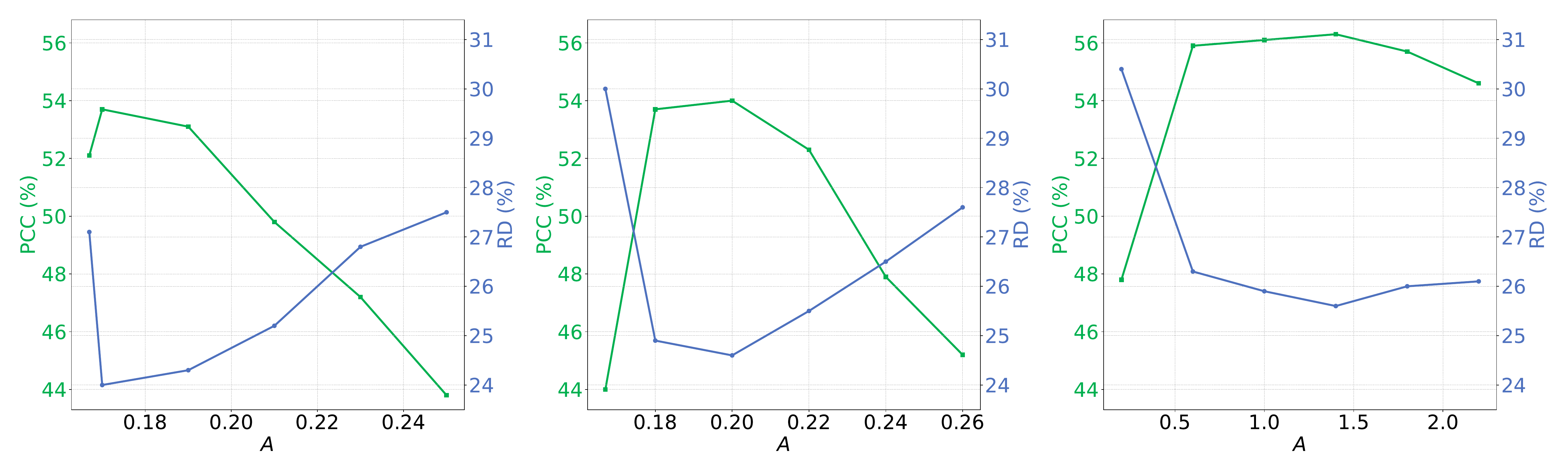}
    \captionof{figure}{The comparison of different PDFs: Gaussian (left), linear (middle), and exponential (right). ``$A$'' is the amplitude in each PDF function (Eqn.~\ref{Eq: g} to Eqn.~\ref{Eq: e}).} 
    \label{Fig: plot_output}
\end{figure*}

\textbf{Comparison against two non-PDF methods.} In Table \ref{Table: demo}, we also show the experimental results of our method under two standard supervising signals, \emph{i.e.}, ``One-hot Label” and ``Label Smoothing ($\epsilon = 0.5$)”. 
In comparison, our \msname~ using PDFs gains significant superiority, \emph{e.g.}, using exponential PDF is better than label smoothing by $+21.3\%$ PCC and $-10.5\%$ RD. This superiority validates our intuition that neighboring replication levels should be continuous and smooth.


\textbf{Comparison between different PDF implementations.} We compare between three different PDF implementations for the proposed \msname~ in Fig. \ref{Fig: plot_output}. 
We observe that: (1) The exponential (convex) function benefits the PCC metric, whereas the Gaussian (concave) function favors the RD metric. The performance of the linear function, which lacks curvature, falls between that of the convex and concave functions. (2) Our method demonstrates robust performance across various distributions, reducing the challenge of selecting an optimal parameter. For example, when using the exponential function, the performance remains high when $A$ ranges from $0.6$ to $1.8$. (3) A model supervised by a smooth PDF outperforms that supervised by a steeper one (also see the corresponding distributions in Fig. \ref{Fig: distribution} of the Appendix). That consists with our intuition again.


\textbf{Our model has good generalizability compared to all other methods.} Because the collection process of the images from some diffusion models (see Appendix \ref{supple: detail}) differs from the process used to build the test set of our D-Rep dataset, it is difficult to label 6 levels for them and the proposed PCC and RD are not suitable. In the Table \ref{Table: plausible}, we consider a quantitative evaluation protocol that measures the average similarity predicted by a model for given $N$ image pairs, which are manually labeled with the highest level. When normalized to a range of $0$ to $1$, a larger value implies better predictions. This setting is practical because, in the real world, most people’s concerns focus on where replication indeed occurs. We manually confirm $100$ such pairs for each diffusion model. We draw three conclusions: (1) Our PDF-Embedding is more generalizable compared to all zero-shot solutions, such as CLIP, GPT4-V, and DINOv2; (2) Our PDF-Embedding still surpasses all other plausible methods trained on the D-Rep dataset in the generalization setting; (3) Compared against testing on SD1.5 (same domain), for the proposed PDF-Embedding, there is no significant performance drop on the generalization setting.

\begin{table}[t]
\caption{The experiments for ``Generalizability to other datasets or diffusion models”. The \colorbox{lightgray!25!white}{gray color} indicates training and testing on the images generated by the same diffusion model.} 
\vspace*{1mm}
\small
  \begin{tabularx}{\hsize}{>{\centering\arraybackslash}p{1.4cm}|>{\centering\arraybackslash}p{2.3cm}|Y|Y|Y|Y|Y|Y|Y}
    \hline
    \multirow{2 }{*}{Class}& \multirow{2}{*}{Method} &\multirow{2}{*}{SD1.5}&Midjo-urney & DAL-L·E 2 & DeepFl-oyd IF & New Bing &\multirow{2}{*}{SDXL} & \multirow{2}{*}{GLIDE} \\ \hline
    \multirow{3 }{*}{Vision-}& SLIP \cite{mu2022slip}  & 0.685 &0.680  & 0.668 &0.710 & 0.688 &0.718& 0.699 \\
    \multirow{3}{*}{language}& BLIP \cite{li2022blip}  & 0.703 &0.674  & 0.673 & 0.696& 0.696 &0.717 &0.689\\
    \multirow{3}{*}{Models}& ZeroVL \cite{cui2022contrastive} & 0.578& 0.581&  0.585 &0.681& 0.589 & 0.677&0.707 \\ 
    & CLIP \cite{radford2021learning}  & 0.646 & 0.665 & 0.694 & 0.728& 0.695 &0.735 &0.727\\ 
    & GPT-4V \cite{openai2023gpt4}  &  0.661 & 0.655 & 0.705 &0.731& 0.732& 0.747 &0.744 \\ 
    \hline
    \multirow{2 }{*}{Self-}& SimCLR \cite{chen2020simple}  & 0.633 & 0.640 & 0.644 &0.656 &0.649 & 0.651& 0.655\\
    \multirow{2 }{*}{supervised}& MAE \cite{he2022masked}   & 0.489 & 0.488 & 0.487 &0.492& 0.487& 0.489 & 0.490\\
    \multirow{2 }{*}{Learning}& SimSiam \cite{chen2021exploring} & 0.572 & 0.611 & 0.619 &0.684 &0.620& 0.645 & 0.683\\
    \multirow{2 }{*}{Models}& MoCov3 \cite{he2020momentum} &  0.585& 0.526 & 0.535 &0.579&0.541& 0.554 & 0.599\\ 
    & DINOv2 \cite{oquab2023dinov2} & 0.766 & 0.529 & 0.593 & 0.723& 0.652 &0.734&0.751 \\ 
    \hline
    \multirow{3 }{*}{Supervised} & EfficientNet \cite{tan2019efficientnet} & 0.116 &0.185&  0.215& 0.241&0.171  &0.210  &0.268 \\
    \multirow{3 }{*}{Pre-trained} & Swin-B \cite{liu2021swin}   &0.334 & 0.387 &0.391  & 0.514 & 0.409& 0.430 & 0.561\\
    \multirow{3 }{*}{Models} & ConvNeXt \cite{liu2022convnet}   & 0.380 &0.429& 0.432 & 0.543 & 0.433&  0.488& 0.580\\ 
    & DeiT-B \cite{touvron2021training}  & 0.386 & 0.478 &0.496  & 0.603& 0.528 &0.525& 0.694\\
    & ResNet-50 \cite{he2016deep}   & 0.362 &  0.436& 0.465 & 0.564&0.450 & 0.522& 0.540\\ 
    \hline
    \multirow{3 }{*}{Current}& ASL \cite{wang2023benchmark}  & 0.183 & 0.231 &0.093& 0.122 & 0.048&  0.049& 0.436\\ 
    \multirow{3 }{*}{ICD}& CNNCL \cite{yokoo2021contrastive}   & 0.201 &0.311 & 0.270 &0.347& 0.279&  0.358& 0.349\\ 
    \multirow{3 }{*}{Models}& SSCD \cite{pizzi2022self}  & 0.116 & 0.181 &0.180 &0.303 & 0.166& 0.239 & 0.266\\
    & EfNet \cite{papadakis2021producing}  & 0.133 &  0.265& 0.267 &0.438& 0.249& 0.340 & 0.349\\
    & BoT \cite{wang2021bag}   & 0.216 &0.345  & 0.346 & 0.477
    &0.338& 0.401 &0.489 \\\hline
    \multirow{3 }{*}{Models}& Enlarging PCC   &\cellcolor{lightgray!25!white}0.598 & 0.510&  0.523& 0.595 & 0.506& 0.554 & 0.592\\ 
    \multirow{3 }{*}{Trained}& Reducing RD  & \cellcolor{lightgray!25!white} 0.795&0.736&  0.736& 0.768 & 0.729& 0.768 & 0.785\\ 
    \multirow{3 }{*}{on \dsname~}& Regression   &\cellcolor{lightgray!25!white}0.750& 0.694 & 0.705 &0.739  &0.704&  0.721& 0.744\\
    & One-hot Label  &\cellcolor{lightgray!25!white}0.630  & 0.376& 0.400 &0.562 &0.500& 0.548 &0.210 \\
    & Label Smoothing   & \cellcolor{lightgray!25!white} 0.712& 0.568 &0.636&  0.628&0.680& 0.676 & 0.548\\
    \hline
    \multirow{3 }{*}{\textbf{Ours}}& Gaussian PDF & \cellcolor{lightgray!25!white} 0.787&0.754& 0.784  & 0.774 &0.774 &  0.780& 0.776\\ 
    & Linear PDF&\cellcolor{lightgray!25!white} 0.822 &0.758  & 0.798 & 0.794& 0.782& 0.794& 0.790\\ 
    & Exponential PDF & \cellcolor{lightgray!25!white}\textbf{0.831} & \textbf{0.814} & \textbf{0.826} &\textbf{0.804} &\textbf{0.802} &\textbf{0.818} & \textbf{0.794}\\ 
    \hline
  \end{tabularx}
  \label{Table: plausible}
\end{table}

\subsection{Simulated Evaluation of Diffusion Models}\label{Sec: real}
\begin{figure*}[t]
    \includegraphics[width=14cm]{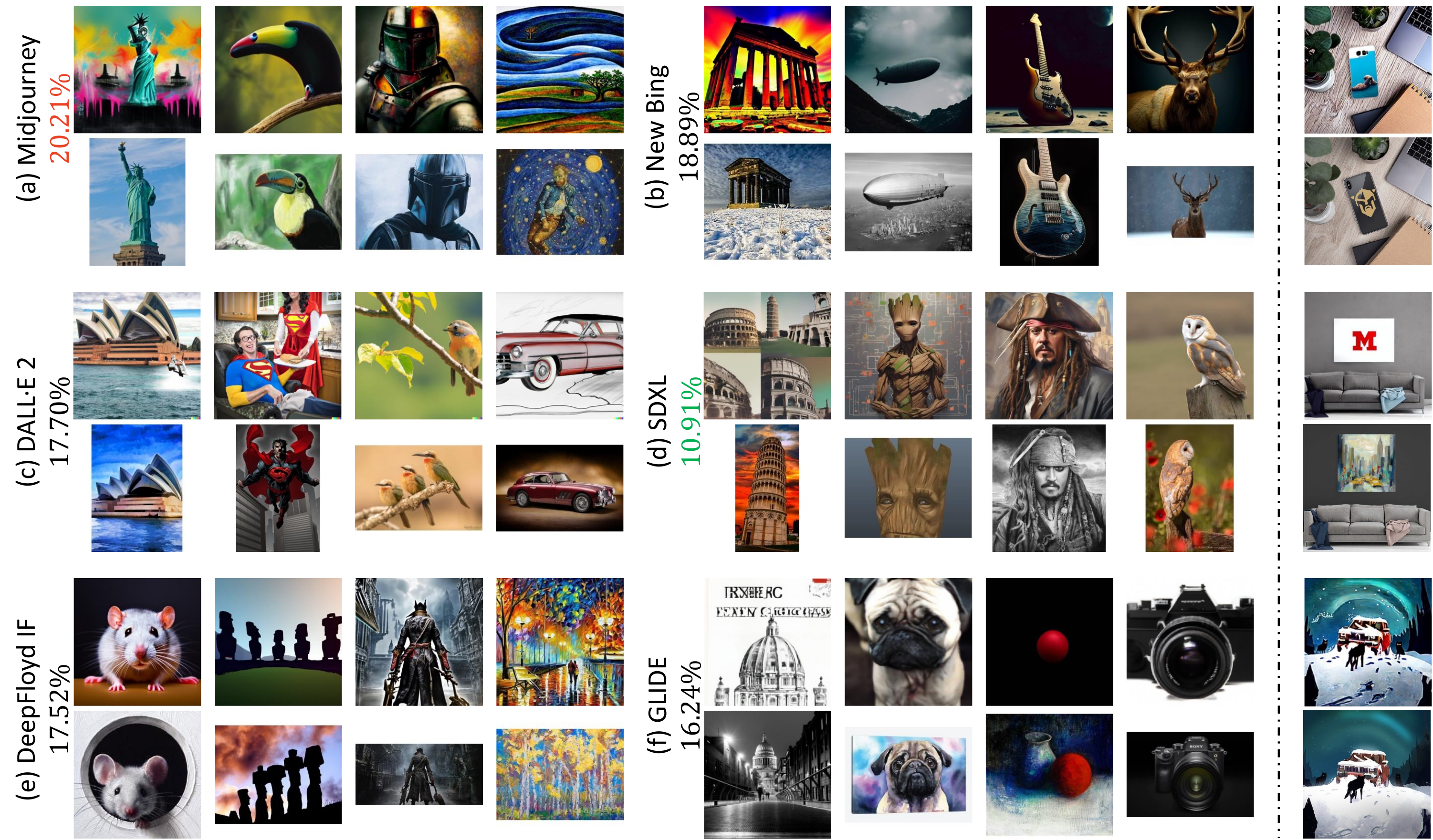}
    \captionof{figure}{Left: Examples of diffusion-based replication fetched by our \msname~. The accompanying percentages indicate the replication ratio of each model. Right: Examples filtered by SSCD \cite{pizzi2022self} in \cite{somepalli2023diffusion}. Compared to them, our results are more diverse: For example, the ``Groot” generated by SDXL includes the whole body, whereas the original one features only the face; and the ``Moai statues” created by DeepFloyd IF are positioned differently compared to the original image. } 
    \label{Fig: famous_all}
\end{figure*}
In this section, we simulate a scenario using our trained \msname~ to evaluate popular diffusion models. We select 6 famous diffusion models, of which three are commercial, and another three are open source (See Section \ref{supple: detail} in the Appendix for more details). We use the LAION-Aesthetics V2 6+ dataset \cite{schuhmann2022laion} as the gallery and investigate whether popular diffusion models replicate it. When assessing the replication ratio of diffusion models, we consider image pairs rated at Level 4 and Level 5 to be replications.

\textbf{Evaluation results.} Visualizations of matched examples and the replication ratios are shown in Fig. \ref{Fig: famous_all} (Left). For more visualizations, please refer to the Appendix (Section \ref{supple: piracy}). We observe that the replication ratios of these diffusion models roughly range between $10\%$ and $20\%$. The most ``aggressive” model is Midjourney \cite{midjourney2022} with a rate of $20.21\%$, whereas the ``conservative” model is SDXL \cite{podell2023sdxl} at $10.91\%$. We also include an analysis of failure cases in the Appendix (Section \ref{App: fail}).\par 

\textbf{Efficiency analysis.} Efficiency is crucial in real-world scenarios. A replication check might slow down the image generation speed of diffusion models. Our \msname~ requires only $2.07 \times 10^{-3}$ seconds for inference and an additional $8.36 \times 10^{-2}$ seconds for matching when comparing a generated image against a reference dataset of $12$ million images using a standard A100 GPU. This time overhead is negligible compared to the time required for generating (several seconds).\par


\textbf{Intuitive comparison with another ICD model.} In \cite{somepalli2023diffusion}, SSCD \cite{pizzi2022self} is used as a feature extractor to identify replication, as illustrated in Fig. \ref{Fig: famous_all} (Right). In comparison, our \msname~ detects a higher number of challenging cases (``hard positives”). Despite visual discrepancies between the generated and original images, replication has indeed occurred.


\section{Conclusion}
This paper investigates a particular and critical Image Copy Detection (ICD) problem: Image Copy Detection for Diffusion Models (\tsname~). 
We introduce the first \tsname~ dataset and propose a strong baseline called ``\msname~”. A distinctive feature of the \dsname~ is its use of replication levels. 
The dataset annotates each replica into 6 different replication levels. The proposed \msname~ first transforms the annotated level into a probability density function (PDF) to smooth the probability. To learn from the PDFs, our \msname~ adopts a set of representative vectors instead of a traditional representative vector. 
We hope this work serves as a valuable resource for research on replication in diffusion models and encourages further research efforts in this area.

\textbf{Disclaimer.}
The model described herein may yield false positive or negative predictions. Consequently, the contents of this paper should not be construed as legal advice.

{
    \small
    \bibliography{neurips_2024}
    \bibliographystyle{unsrt}
}


\appendix

\clearpage

\section{More examples of the \dsname~ Dataset}
\label{supple: more_example}
We show more example image pairs for each level in Fig. \ref{Fig: label_5} to Fig. \ref{Fig: label_0}.


\section{The Instantiations of PDFs}
\label{supple: pdf}
This section presents examples of PDFs derived from replication levels, focusing on three primary functions: Gaussian, linear, and exponential for our calculations, visualization, and analysis. Within the area close to the normalized level $p^l$, denoted as $\delta$, the Gaussian function curves downwards making it concave, the linear function is straight with no curvature, and the exponential function curves upwards making it convex. These characteristics indicate the rate at which they deviate from their peak value: the Gaussian function changes slowly in the $\delta$ area, the linear function changes at a steady rate, and the exponential function changes rapidly in the $\delta$ area. A fast rate of change suggests the network learns from a sharp distribution, while a slow rate implies learning from a smooth distribution.

\textbf{Gaussian function.} Its general formulation is 
\begin{equation}
g(x \mid A, \mu, \sigma)=A \cdot \exp \left(-\frac{(x-\mu)^2}{2 \cdot \sigma^2}\right),
\end{equation}
where $A > 0$ is the amplitude (the height of the peak), $\mu \in \left[ 0,1\right]  $ is the mean or the center, and $\sigma>0$ is the standard deviation. To satisfy the requirements of a PDF in Section \ref{Sec: req}, the following must hold:
\begin{equation}\label{Eq: guass}
\begin{gathered}\int^{1}_{0} \left( A\cdot \exp \left( -\frac{(x-\mu )^{2}}{2\cdot \sigma^{2} } \right)  \right)  dx=1,\\ A\cdot \exp \left( -\frac{(x-\mu )^{2}}{2\cdot \sigma^{2} } \right)  \geq 0,\\ A\cdot \exp \left( -\frac{(x-\mu )^{2}}{2\cdot \sigma^{2} } \right)  \leq A\cdot \exp \left( -\frac{(p^l-\mu )^{2}}{2\cdot \sigma^{2} } \right).  \end{gathered}
\end{equation}
From Eqn. \ref{Eq: guass}, we have:
\begin{equation}
\mu \  =\  p^l.
\end{equation}
In practice, with $x\in \left\{ 0,0.2,0.4,0.6,0.8,1\right\}  $ being discrete, the equations become:

\begin{equation}
\begin{gathered}\sum_{x\in \{ 0,0.2,0.4,0.6,0.8,1\} } \left( A\cdot \exp \left( -\frac{(x-p^l)^{2}}{2\cdot \sigma^{2} } \right)  \right)  =1,\\ A\cdot \exp \left( -\frac{(x-p^l)^{2}}{2\cdot \sigma^{2} } \right)  \geq 0.\end{gathered} 
\end{equation}
Given a specific normalized level $p^l$ and varying $A$, $g(x \mid A, \mu, \sigma)$ values are computed for different $x$ using numerical approaches. The resulting distributions are visualized in Fig. \ref{Fig: distribution} (top).\par 

Finally we prove that \textbf{\textit{$g(x \mid A, \mu, \sigma)$ is concave for $x$ in the interval $[\mu-\sigma, \mu+\sigma]$. This means that in the interested region $\delta$ (near the normalized level $p^l$), its rate of change is slow and increases as $x$ diverges from $\mu$.}}\par

\textit{Proof: }
Given the function:
\begin{equation}
g(x \mid A, \mu, \sigma)=A \exp \left(-\frac{(x-\mu)^2}{2 \sigma^2}\right),
\end{equation}
we find the first derivative of $g$ with respect to $x$ :
\begin{equation}
g^{\prime}(x)=\frac{d}{d x}\left[A \exp \left(-\frac{(x-\mu)^2}{2 \sigma^2}\right)\right].
\end{equation}
Using the chain rule, we have:
\begin{equation}
g^{\prime}(x)=A \exp \left(-\frac{(x-\mu)^2}{2 \sigma^2}\right) \times \frac{d}{d x}\left[-\frac{(x-\mu)^2}{2 \sigma^2}\right].
\end{equation}
This gives:
\begin{equation}
g^{\prime}(x)=-A \exp \left(-\frac{(x-\mu)^2}{2 \sigma^2}\right) \times \frac{(x-\mu)}{\sigma^2}.
\end{equation}
Next, to find the second derivative, differentiate $g^{\prime}(x)$ with respect to $x$ :
\begin{equation}
g^{\prime \prime}(x)=\frac{d}{d x}\left[-A \exp \left(-\frac{(x-\mu)^2}{2 \sigma^2}\right) \times \frac{(x-\mu)}{\sigma^2}\right].
\end{equation}
Using product rule and simplifying, the result would be:
\begin{equation}
g^{\prime \prime}(x)=A \exp \left(-\frac{(x-\mu)^2}{2 \sigma^2}\right) \times\left[\frac{(x-\mu)^2}{\sigma^4}-\frac{1}{\sigma^2}\right].
\end{equation}
When $g^{\prime \prime}(x)<0$, we have:
\begin{equation}
x\in \left[ \mu -\sigma ,\mu +\sigma \right].
\end{equation}

That proves $g(x \mid A, \mu, \sigma)$ is concave in the $\delta$ area, and thus its rate of change is slow and increases as $x$ diverges from $\mu$.

\textbf{Linear function.} Its general formulation is
\begin{equation}
g(x \mid A, \mu, \beta) = A - \beta \cdot |x-\mu|,
\end{equation}
where $A>0$ denotes the maximum value of the function, $\mu \in \left[ 0,1\right]  $ is the point where the function is symmetric, and $\beta>0$ determines the function's slope. To satisfy the requirements of a PDF in Section \ref{Sec: req}, the following must hold:
\begin{equation}\label{Eq: linear}
\begin{gathered}\int^{1}_{0} \left( A-\beta \cdot |x-\mu |\right)  dx=1,\\ A-\beta \cdot |x-\mu |\geq 0,\\ A-\beta \cdot |x-\mu |\leq A-\beta \cdot |p^l-\mu |.\end{gathered} 
\end{equation}
From Eqn. \ref{Eq: linear}, we have:
\begin{equation}
\mu \  =\  p^l.
\end{equation}
In practice, with $x\in \left\{ 0,0.2,0.4,0.6,0.8,1\right\}  $ being discrete, the equations become:
\begin{equation}
\begin{gathered}\sum_{x\in \{ 0,0.2,0.4,0.6,0.8,1\} } (A-\beta \cdot |x-p^l|)=1,\\ A-\beta \cdot |x-p^l|\geq 0.\end{gathered} 
\end{equation}
Given a specific normalized level $p^l$ and varying $A$, $g(x \mid A, \mu, \beta)$ values are computed for different $x$. The resulting distributions are visualized in Fig. \ref{Fig: distribution} (middle).\par 

Finally, we prove that \textbf{\textit{$g(x \mid A, \mu, \beta)$ has no curvature, and thus its rate of change is consistent regardless of the value of $x$.}} \par 
\textit{Proof:} Given the function:
\begin{equation}
g(x \mid A, \mu, \beta)=A-\beta \cdot|x-\mu|,
\end{equation}
we will differentiate this function based on the absolute value, which will result in two cases for the derivatives based on the sign of $(x-\mu)$.\par 
Case 1: $x > \mu$: 
In this case, $|x-\mu|=x-\mu$. So, $g(x \mid A, \mu, \beta)=A-\beta \cdot(x-\mu)$. \par 
First derivative $g^{\prime}(x)$ :
\begin{equation}
g^{\prime}(x)=\frac{d}{d x}(A-\beta \cdot(x-\mu)) = -\beta.
\end{equation}

Second derivative $g^{\prime \prime}(x)$ :
\begin{equation}
g^{\prime \prime}(x)=\frac{d}{d x}(-\beta) = 0.
\end{equation}
\par 
Case 2: $x<\mu$: 
In this case, $|x-\mu|=\mu-x$. So, $g(x \mid A, \mu, \beta)=A-\beta \cdot(\mu-x)$.\par 
First derivative $g^{\prime}(x)$ :
\begin{equation}
 g^{\prime}(x)=\frac{d}{d x}(A-\beta \cdot(\mu-x)) = \beta.
\end{equation}

Second derivative $g^{\prime \prime}(x)$ :
\begin{equation}
g^{\prime \prime}(x)=\frac{d}{d x}(\beta) =0.
\end{equation}

When the second derivative is constantly $0$, it means the function has no curvature and its rate of change is constant at every point.\par

\textbf{Exponential function.} Its general formulation is 

\begin{equation}
g(x \mid A, \mu, \lambda)=A \cdot \lambda \cdot \exp (-\lambda \cdot|x-\mu|),
\end{equation}
where $A>0$ denotes the intensity of the function, $\mu \in \left[ 0,1\right]  $ is the point where the function is symmetric, and $\lambda>0$ determines the spread or width of the function. To satisfy the requirements of a PDF in Section \ref{Sec: req}, the following must hold:
\begin{equation}\label{Eq: exp}
\begin{gathered}\int^{1}_{0} \left( A\cdot \lambda \cdot \exp (-\lambda \cdot |x-\mu |)\right)  dx=1,\\ A\cdot \lambda \cdot \exp (-\lambda \cdot |x-\mu |)\geq 0,\\ A\cdot \lambda \cdot \exp (-\lambda \cdot |x-\mu |)\leq A\cdot \lambda \cdot \exp (-\lambda \cdot |p^l-\mu |).\end{gathered} 
\end{equation}
From Eqn. \ref{Eq: exp}, we have:
\begin{equation}
\mu \  =\  p^l.
\end{equation}
In practice, with $x\in \left\{ 0,0.2,0.4,0.6,0.8,1\right\}  $ being discrete, the equations become:

\begin{equation}
\begin{gathered}\sum_{x\in \{ 0,0.2,0.4,0.6,0.8,1\} } (A\cdot \lambda \cdot \exp (-\lambda \cdot |x-p^l|))=1,\\ A\cdot \lambda \cdot \exp (-\lambda \cdot |x-p^l|)\geq 0.\end{gathered}
\end{equation}
Given a specific normalized level $p^l$ and varying $A$, $g(x \mid A, \mu, \lambda)$ values are computed for different $x$ using numerical approaches. The resulting distributions are visualized in Fig. \ref{Fig: distribution} (bottom).\par 
Finally, we prove that  \textbf{\textit{$g(x \mid A, \mu, \lambda)$ is convex: its rate of change is rapid in the $\delta$ area and decreases as $x$ moves diverges from $\mu$.}} \par 
\textit{Proof:} Given the function:
\begin{equation}
g(x \mid A, \mu, \lambda)=A \cdot \lambda \cdot \exp (-\lambda \cdot|x-\mu|),
\end{equation}
we find the first and second derivatives with respect to $x$. This function involves an absolute value, which will create two cases for the derivatives based on the sign of $(x-\mu)$.
Case 1: $x > \mu$: In this case, $|x-\mu|=x-\mu$. So,
\begin{equation}
g(x \mid A, \mu, \lambda)=A \cdot \lambda \cdot \exp (-\lambda \cdot(x-\mu)).
\end{equation}
First derivative $g^{\prime}(x)$ :
\begin{equation}
\begin{aligned}
& g^{\prime}(x)=A \cdot \lambda \cdot \frac{d}{d x} \exp (-\lambda \cdot(x-\mu)), \\
& g^{\prime}(x)=-A \cdot \lambda^2 \cdot \exp (-\lambda \cdot(x-\mu)).
\end{aligned}
\end{equation}
Second derivative $g^{\prime \prime}(x)$ :
\begin{equation}
\begin{aligned}
& g^{\prime \prime}(x)=-A \cdot \lambda^2 \cdot \frac{d}{d x} \exp (-\lambda \cdot(x-\mu)), \\
& g^{\prime \prime}(x)=A \cdot \lambda^3 \cdot \exp (-\lambda \cdot(x-\mu))>0.
\end{aligned}
\end{equation}
Case 2: $x<\mu$: In this case, $|x-\mu|=\mu-x$. So,
\begin{equation}
g(x \mid A, \mu, \lambda)=A \cdot \lambda \cdot \exp (-\lambda \cdot(\mu-x)).
\end{equation}
First derivative $g^{\prime}(x)$ :
\begin{equation}
\begin{aligned}
& g^{\prime}(x)=A \cdot \lambda \cdot \frac{d}{d x} \exp (-\lambda \cdot(\mu-x)), \\
& g^{\prime}(x)=A \cdot \lambda^2 \cdot \exp (-\lambda \cdot(\mu-x)).
\end{aligned}
\end{equation}
Second derivative $g^{\prime \prime}(x)$ :
\begin{equation}
\begin{aligned}
& g^{\prime \prime}(x)=A \cdot \lambda^2 \cdot \frac{d}{d x} \exp (-\lambda \cdot(\mu-x)), \\
& g^{\prime \prime}(x)=A \cdot \lambda^3 \cdot \exp (-\lambda \cdot(\mu-x))>0.
\end{aligned}
\end{equation}
When the second derivative is bigger than $0$, it means the function is convex: its rate of change is rapid in the $\delta$ area and decreases as $x$ moves diverge from $\mu$.

\begin{figure}[t]
\centering
    \includegraphics[width=8cm]{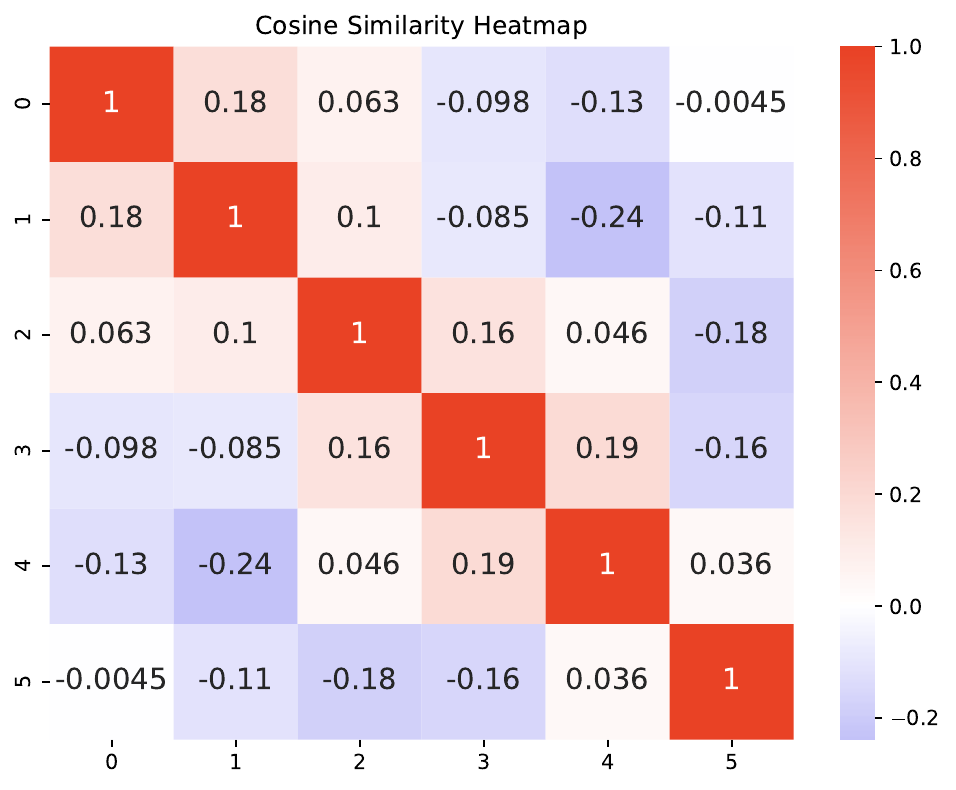}
    \captionof{figure}{The cosine similarity heatmap of the learned vectors.} 

    \label{Fig: heatmap}
\end{figure}

\section{The Visualization of What the Network Learns}\label{supple: learn}
To gain insights into what the network has learned, we offer two visualization methods. First, we present the cosine similarity heatmap of the learned $\mathbf{C}^0=\left[\mathbf{c}_0^0, \mathbf{c}_1^0, \ldots, \mathbf{c}_N^0\right]$ (refer to Fig. \ref{Fig: heatmap}). Second, we show the distribution changes of image pairs throughout the training process. The final epoch's distribution can be seen in Fig. \ref{Fig: learnt_dist}, while the entire training process is depicted in the attached videos.\par 
From the heatmap, we conclude that: (1) The cosine similarity between different vectors is very low. This demonstrates that the learned vectors are linearly independent.
(2) Neighboring vectors exhibit a relatively high cosine similarity. This is consistent with the expectation, as they correspond to similar replication levels.\par 
From the observed changes in the distributions, we note that: (1) While the distribution initially starts as a uniform distribution or peaks at an incorrect level, the network, after training, eventually produces an appropriate and accurate distribution for each image pair. (2) For instance, when supervised by the Gaussian distribution, the network, as expected, produces a final distribution that, though not perfect, closely imitates this.


\section{Implement GPT-4V Turbo on our \dsname~ Test Dataset}\label{supple: gpt}
This section details implementing GPT-4V Turbo on our \dsname~ test dataset. GPT-4V Turbo, which has been online since November 6, 2023, is the latest and most powerful large multimodal model developed by OpenAI. Because it cannot be regarded as a feature extractor, we directly prompt it with two images and one instruction:
\begin{mdframed}
\textit{Give you one pair of images; please give the similarity of the second one to the first one. Diffusion Models generate the second one, while the first one is the original one. Please only reply with one similarity from $0-1$; no other words are needed. Please understand the images by yourself.}
\end{mdframed}
Given these prompts, GPT-4V Turbo returns a similarity ranging from $0$ to $1$. Using the official API, we ask the GPT-4V Turbo to determine all similarities between the image pairs in the \dsname~ test dataset. Note that the computational cost of employing GPT-4V Turbo in practical applications is prohibitively high. Specifically, to compare an image against an image database containing one million images, the API must be called one million times, incurring a cost of approximately $\$7,800$.

\section{The Similarities Predicted by Other Models}\label{supple: similarities}
In Fig. \ref{Fig: sim}, we show the similarities predicted by six selected models (two vision-language models, two current ICD models, and two others). We conclude that: (1) CLIP \cite{radford2021learning} tends to assign higher similarities, which deteriorates its performance on image pairs with low levels, leading to many false positive predictions; (2) GPT-4V Turbo \cite{openai2023gpt4} and DINOv2 \cite{oquab2023dinov2} can produce both high and low predictions, but its performance does not match ours. (3) The prediction ranges of ResNet-50 \cite{he2016deep} are relatively narrow, indicating its inability to distinguish image pairs with varying levels effectively. (4) Current ICD models, including SSCD \cite{pizzi2022self} and BoT \cite{wang2021bag}, consistently produce low predictions. This is because they are trained for invariance to image transformations (resulting in high similarities for pirated content produced by transformations) and cannot handle replication generated by diffusion models.

\section{The Details of Six Diffusion Models}\label{supple: detail}
This section provides details on the evaluation sources for three commercial and three open-source diffusion models.\par 
\textbf{Midjourney} \cite{midjourney2022} was developed by the Midjourney AI company. We utilized a \href{https://huggingface.co/datasets/nateraw/midjourney-texttoimage-new}{dataset} scraped by Succinctly AI under the cc0-1.0 license. This dataset comprises $249,734$ images generated by real users from a public Discord server.\par

\textbf{New Bing} \cite{new_bing}, also known as Bing Image Creator, represents Microsoft's latest text-to-image technique. We utilized the \href{https://github.com/acheong08/BingImageCreator}{repository} under the Unlicense to generate $216,957$ images. These images were produced using randomly selected prompts from DiffusionDB \cite{wang-etal-2023-diffusiondb}. \par
\textbf{DALLE·2} \cite{ramesh2022hierarchical} is a creation of OpenAI. We downloaded all generated images directly from this \href{https://dalle2.gallery/#search}{website}, resulting in a dataset containing $50,904$ images. We have obtained permission from the website's owners to use the images for research purposes.\par 
We downloaded and deployed three open-source diffusion models, including \textbf{SDXL} \cite{podell2023sdxl}, \textbf{DeepFloyd IF} \cite{deep_floyd_IF}, and \textbf{GLIDE} \cite{nichol2022glide}. These models were set up on a local server equipped with 8 A100 GPUs. Distributing on them, we generated $1,819,792$ images with prompts from DiffusionDB \cite{wang-etal-2023-diffusiondb}.\par

\section{More Replication Examples}\label{supple: piracy}
We provide more replication examples by diffusion models in Fig. \ref{Fig: midall} to Fig. \ref{Fig: glideall}.

\section{Failure Cases}\label{App: fail}
\begin{figure}
    \includegraphics[width=14cm]{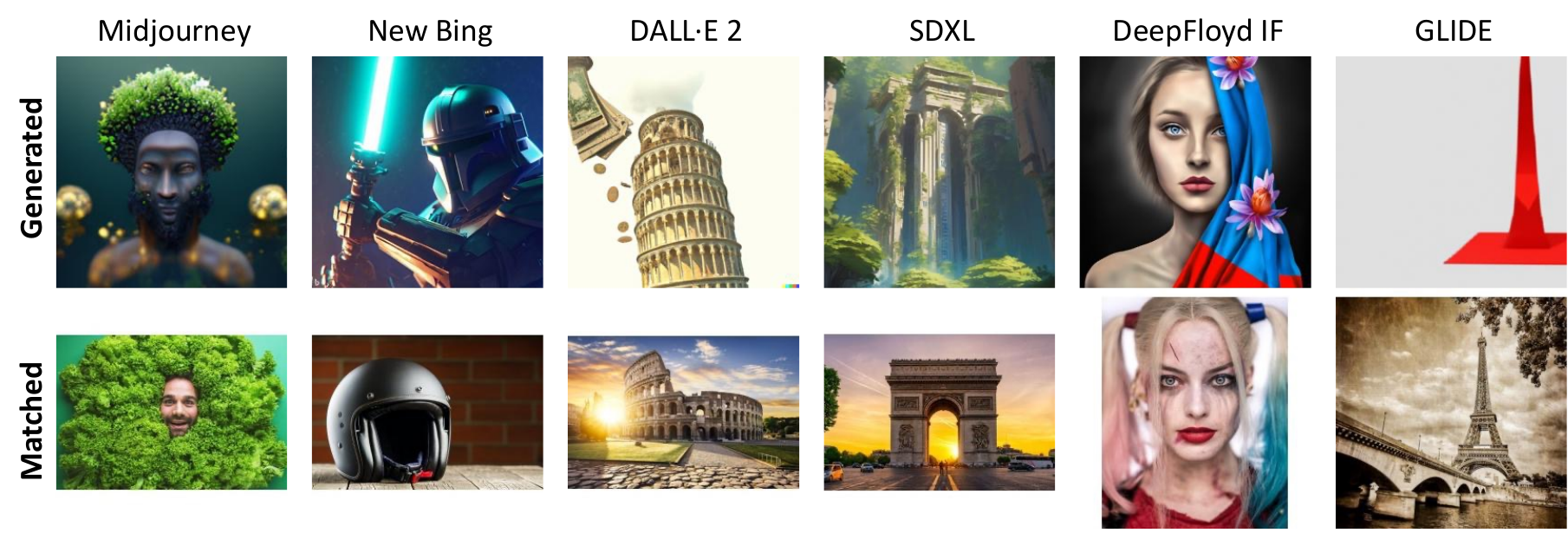}
    \centering
    \captionof{figure}{The failure cases of our detection method. We show one example for each diffusion model.} 
    \label{Fig: failure}
\end{figure}

As shown in Fig. \ref{Fig: failure}, we identify two primary failure cases. The first type of failure occurs when the generated images replicate only common elements without constituting replicated content.  For instance, elements like grass (Midjourney), helmets (New Bing), and buildings (DALL·E 2) appear frequently but do not indicate actual replication of content. The second type of failure arises when two images share high-level semantic similarity despite having no replicated content. An example can be seen in the image pairs where themes, styles, or concepts are similar, such as the presence of iconic structures (SDXL and GLIDE) or stylized portraits (DeepFloyd IF), even if the specific content is not replicated. Understanding these failure modes is crucial for improving the accuracy and robustness of our detection methods in the future.



\newpage


\begin{figure}[H]
\centering
\vspace{6mm}
    \includegraphics[width=15cm]{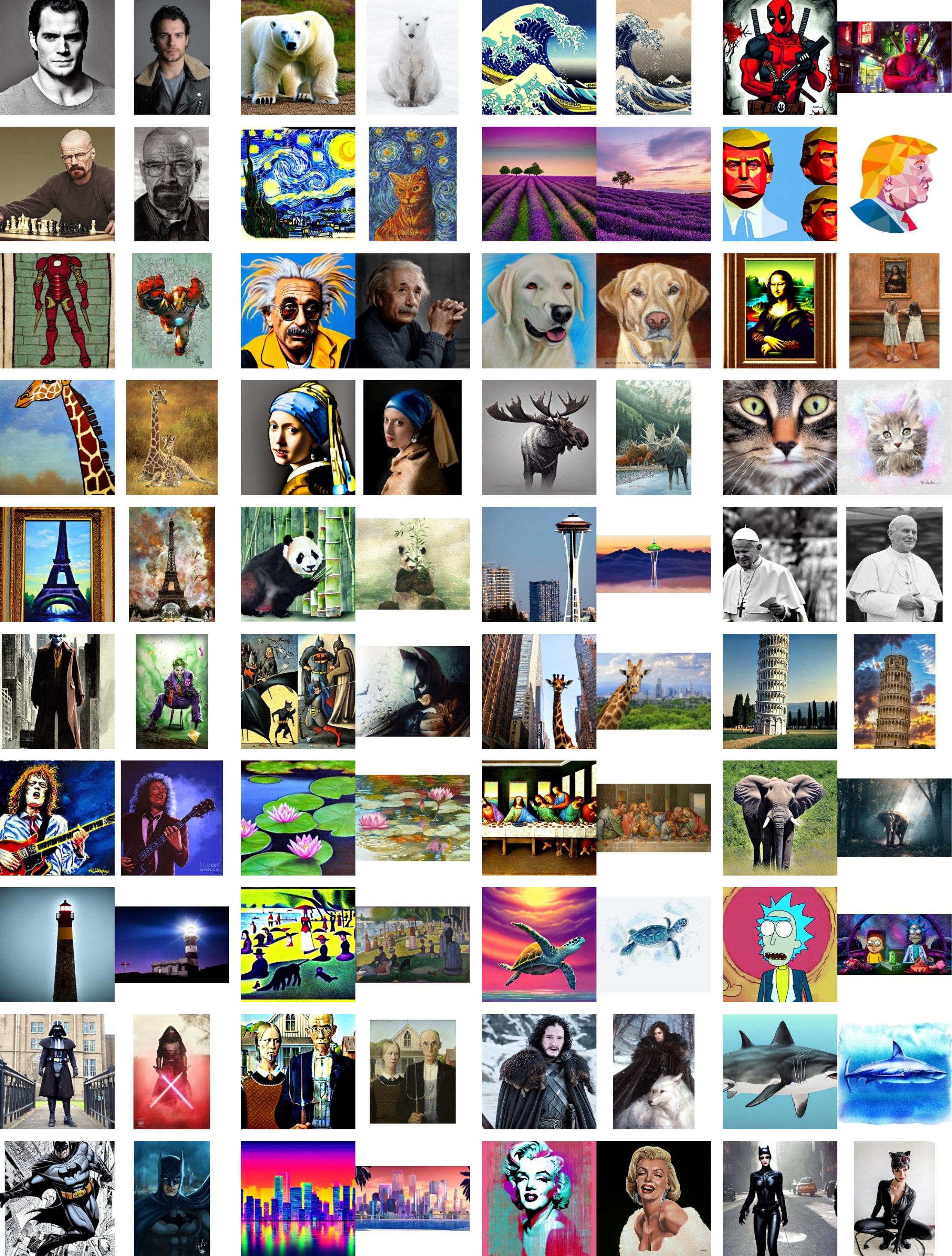}
    \vspace{3mm}
    \captionof{figure}{The example image pairs with level $5$.} 

    \label{Fig: label_5}
\end{figure}

\begin{figure}[H]
\centering
\vspace{6mm}
    \includegraphics[width=15cm]{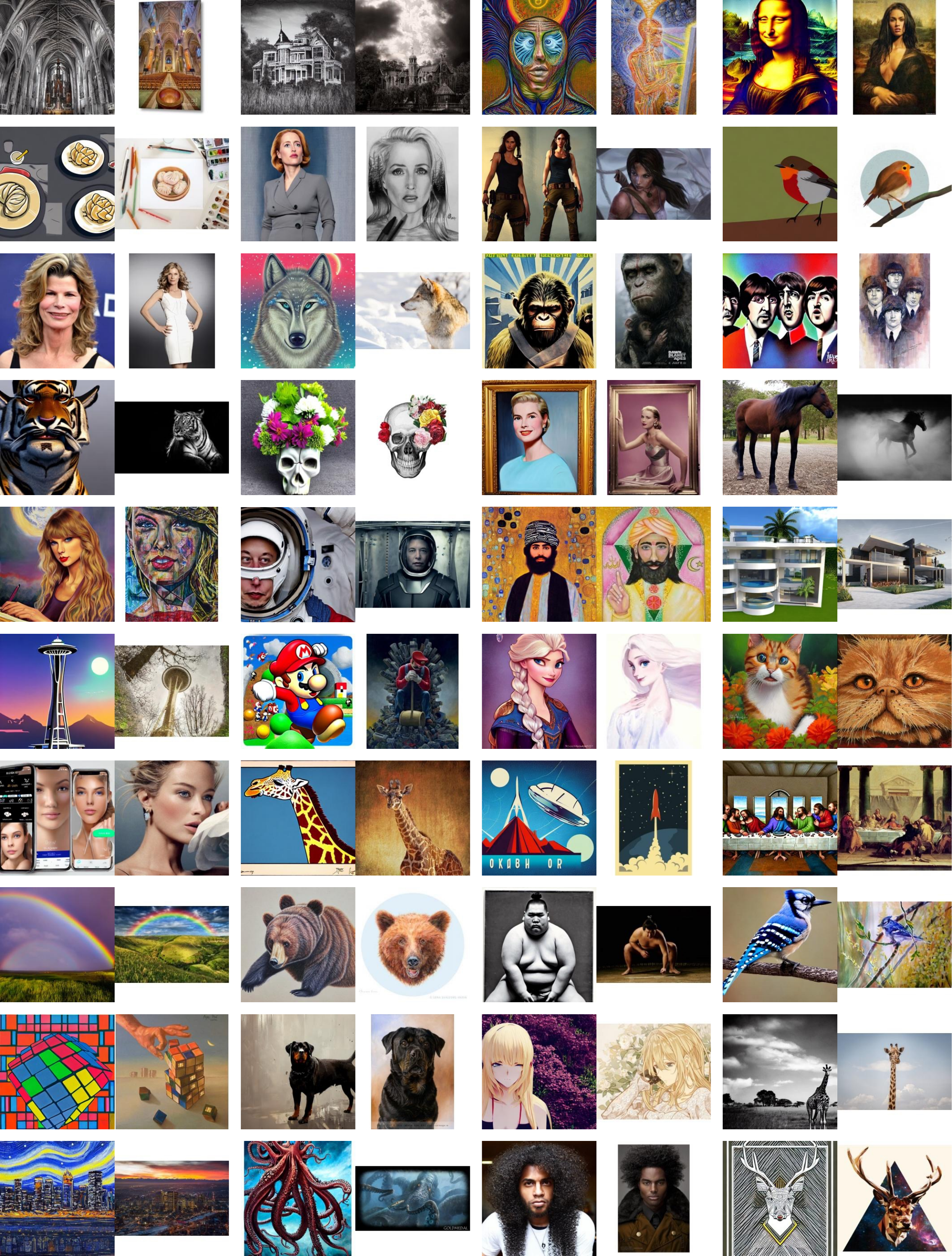}
    \vspace{3mm}
    \captionof{figure}{The example image pairs with level $4$.} 

    \label{Fig: label_4}
\end{figure}

\begin{figure}[H]
\centering
\vspace{6mm}
    \includegraphics[width=15cm]{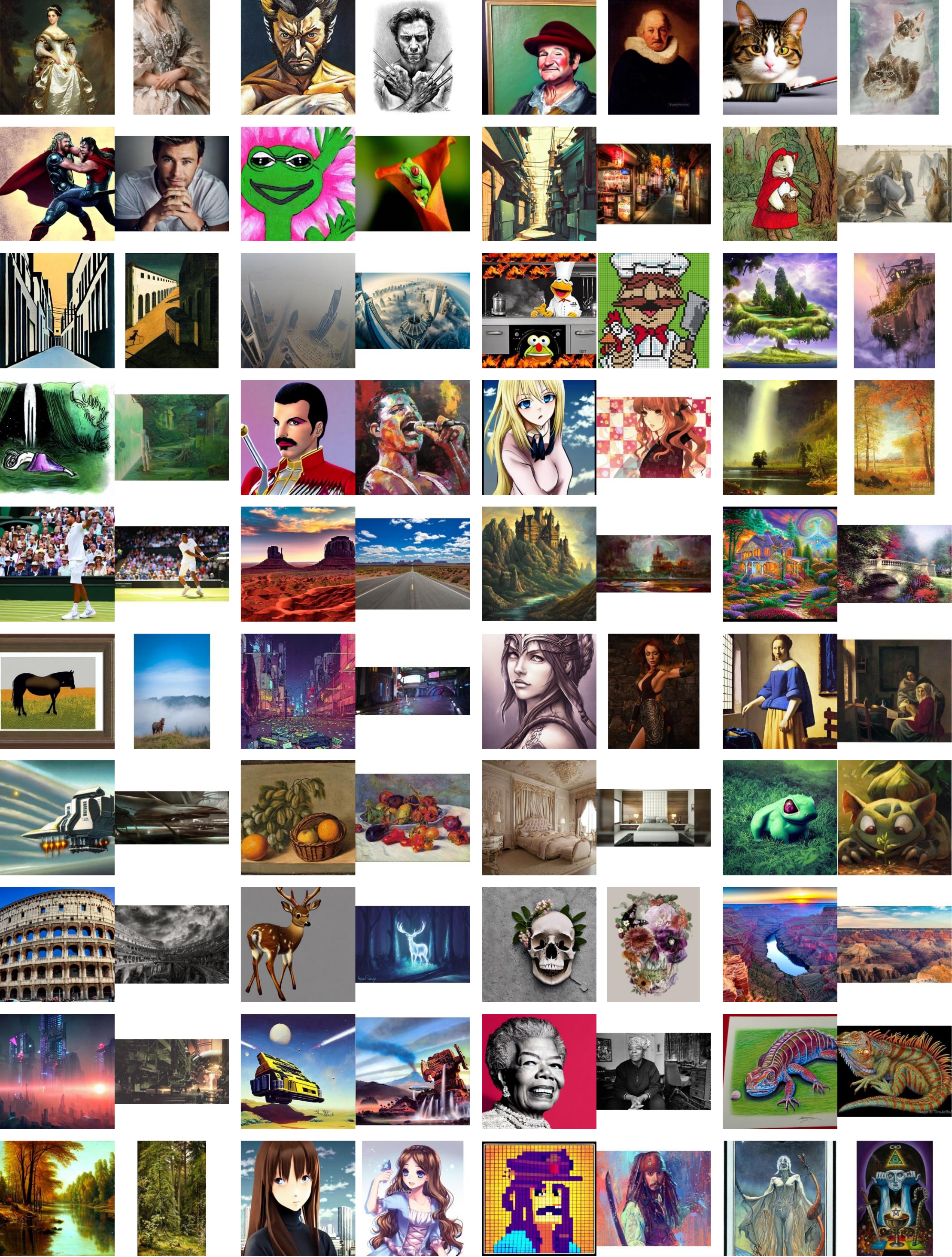}
    \vspace{3mm}
    \captionof{figure}{The example image pairs with level $3$.} 

    \label{Fig: label_3}
\end{figure}

\begin{figure}[H]
\centering
\vspace{6mm}
    \includegraphics[width=15cm]{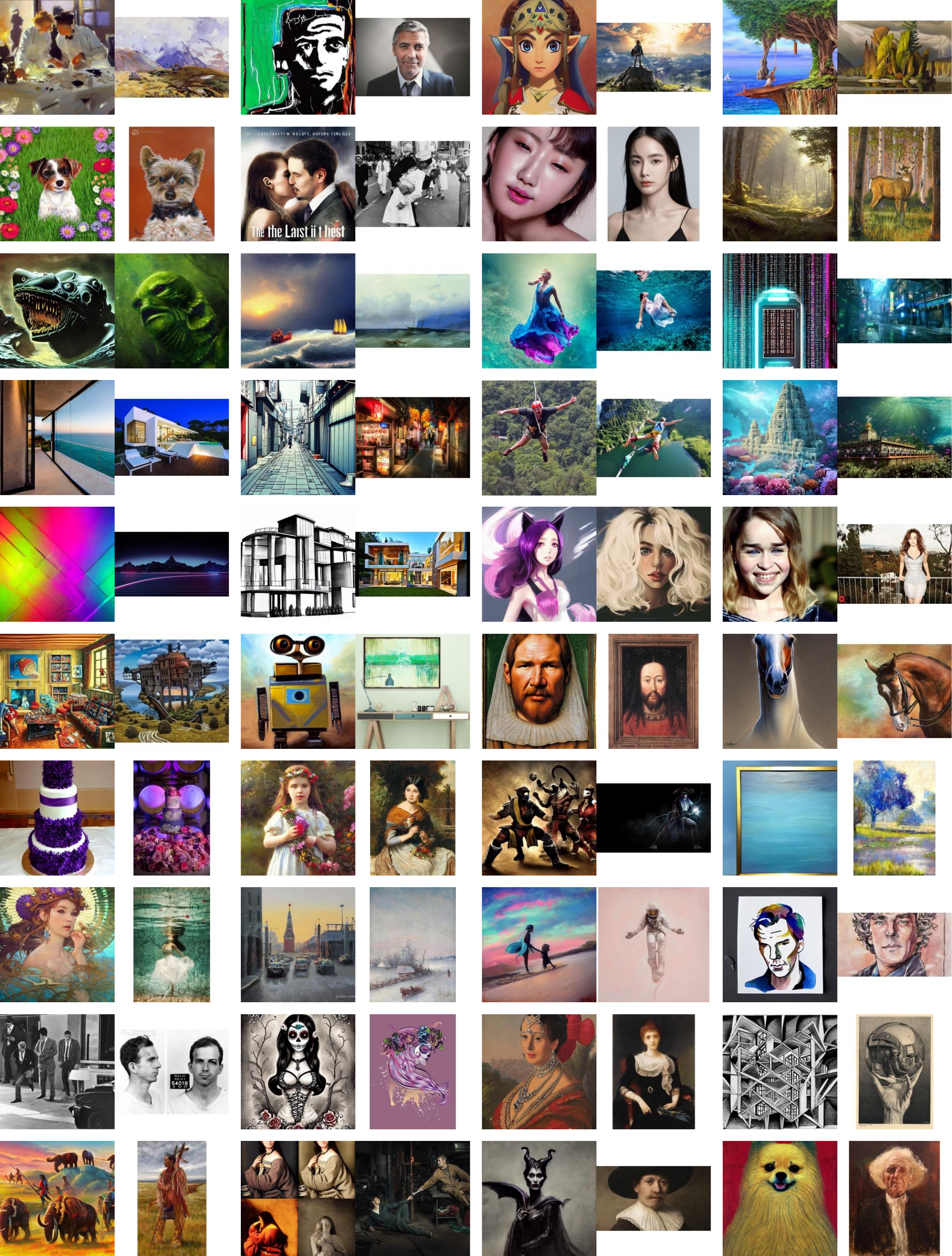}
    \vspace{3mm}
    \captionof{figure}{The example image pairs with level $2$.} 

    \label{Fig: label_2}
\end{figure}

\begin{figure}[H]
\centering
\vspace{6mm}
    \includegraphics[width=15cm]{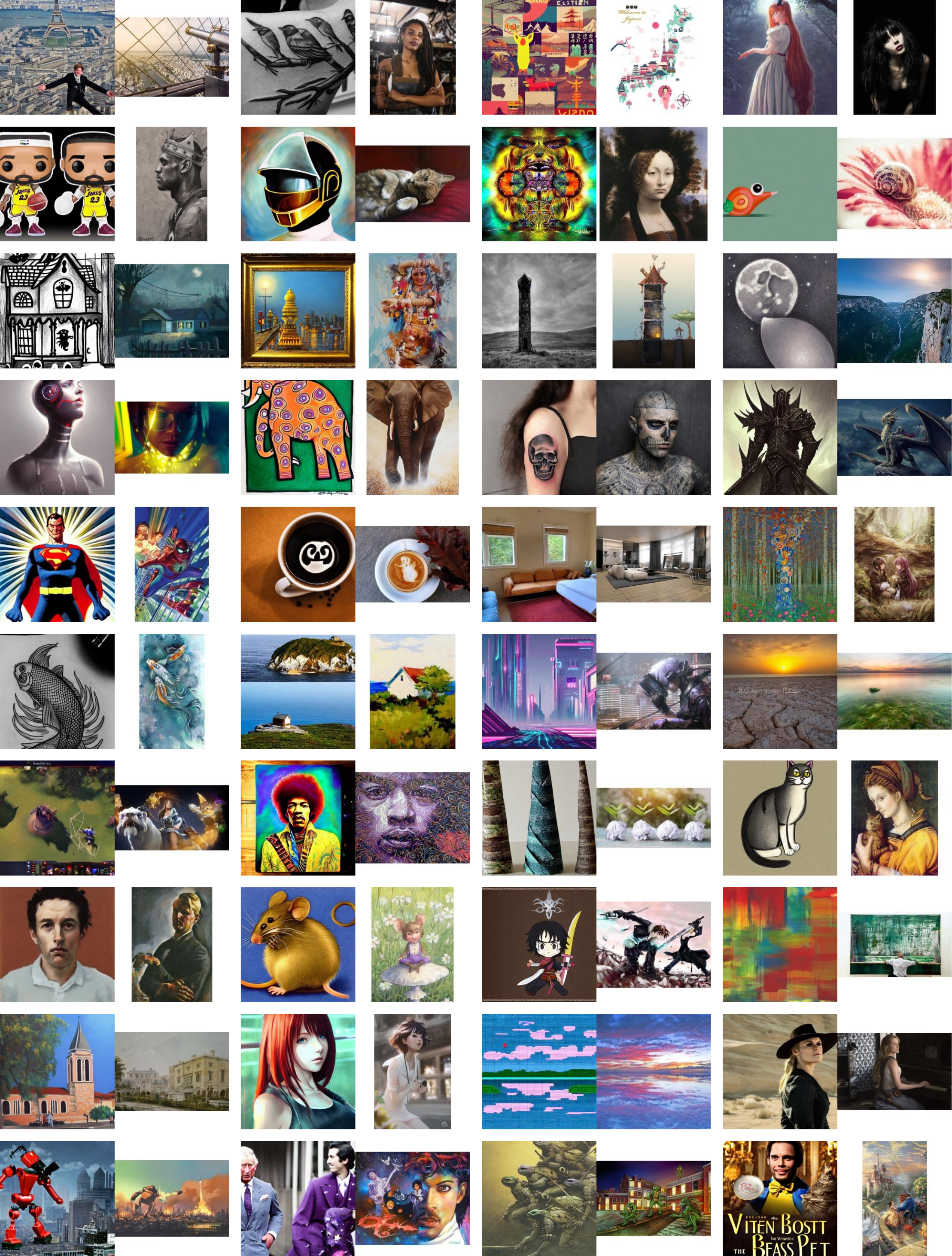}
    \vspace{3mm}
    \captionof{figure}{The example image pairs with level $1$.} 

    \label{Fig: label_1}
\end{figure}

\begin{figure}[H]
\centering
\vspace{6mm}
    \includegraphics[width=15cm]{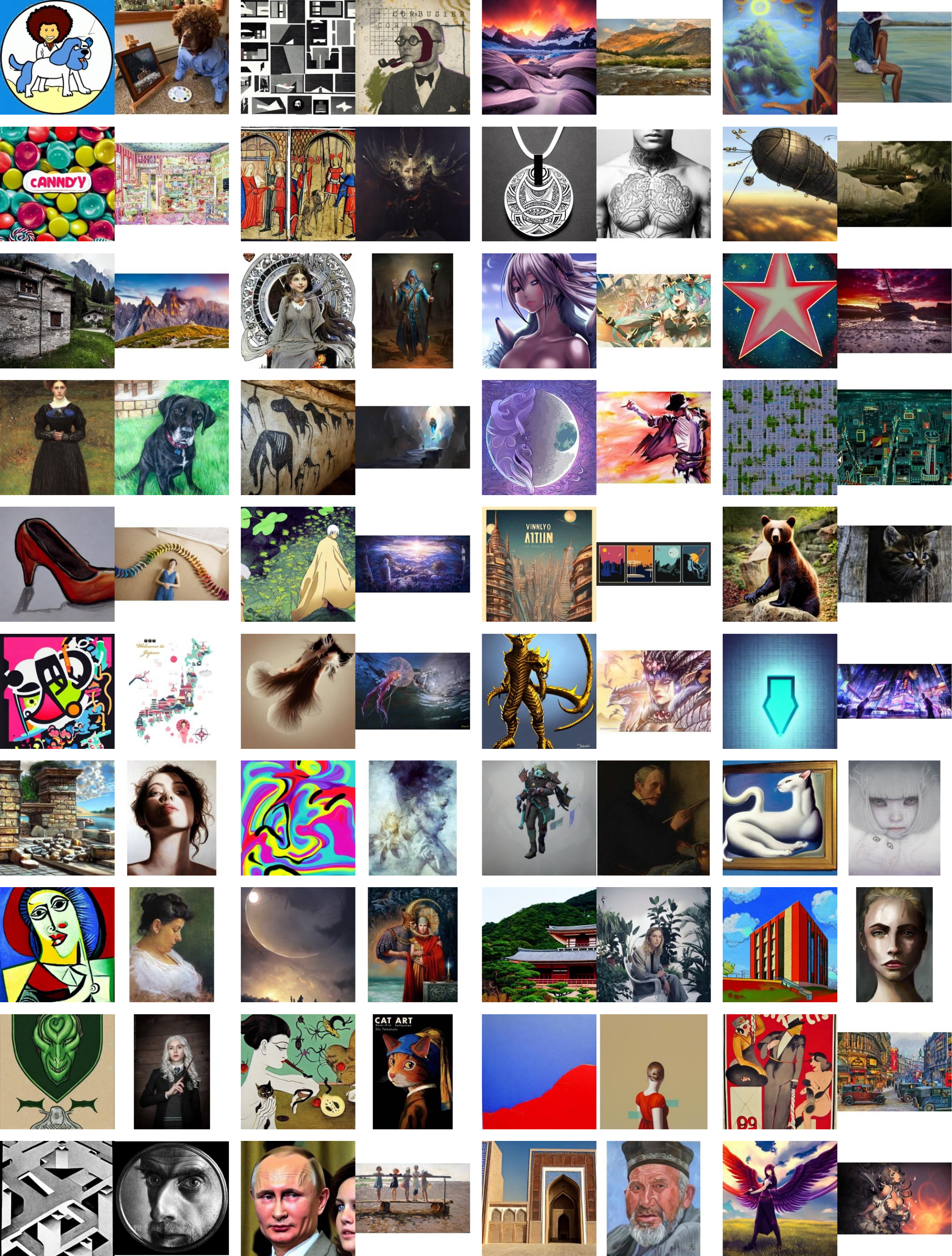}
    \vspace{3mm}
    \captionof{figure}{The example image pairs with level $0$.} 

    \label{Fig: label_0}
\end{figure}

\begin{figure}[H]
\centering
\vspace{6mm}
    \includegraphics[width=14cm]{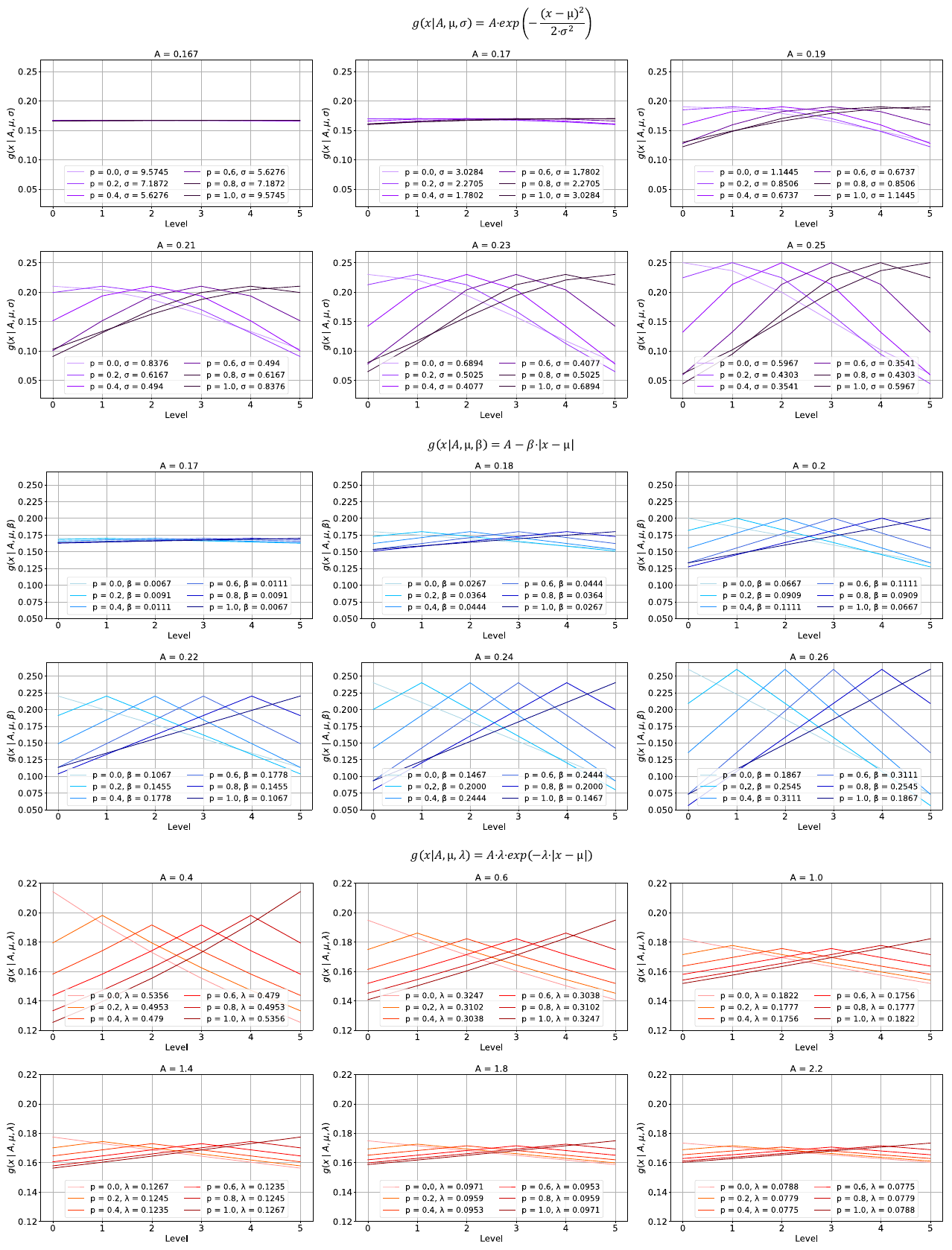}
    \vspace{3mm}
    \captionof{figure}{The distributions converted from replication levels. We use Gaussian, linear, and exponential functions as the representative demonstrations.} 

    \label{Fig: distribution}
\end{figure}

\begin{figure}[H]
\centering
\vspace{6mm}
    \includegraphics[width=14cm]{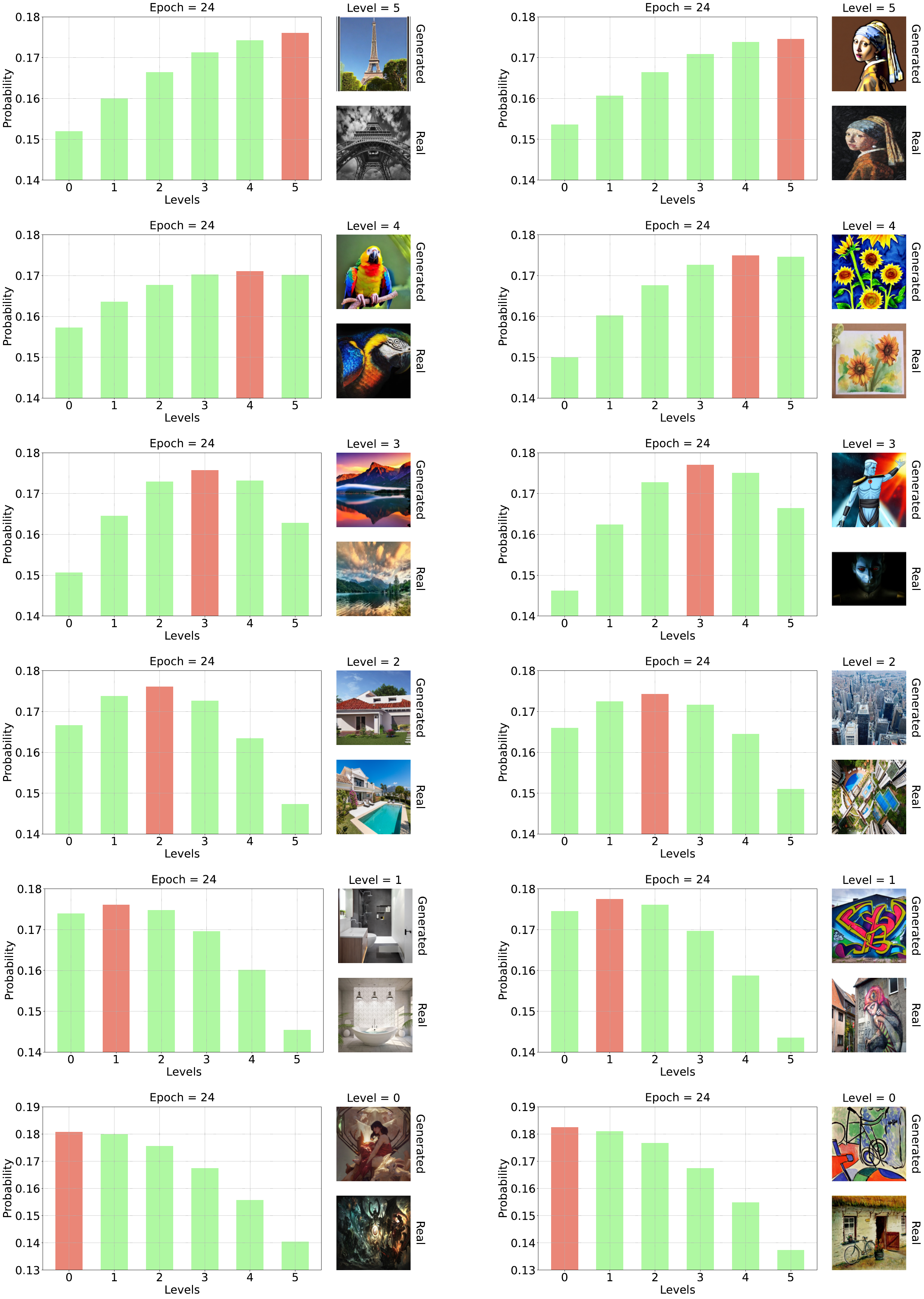}
    \vspace{3mm}
    \captionof{figure}{The learned distributions of different image pairs. Please see the attached videos for the distribution changes in the whole training process.} 

    \label{Fig: learnt_dist}
\end{figure}

\begin{figure}[H]
    \begin{minipage}{0.13\linewidth}
    \centering
    \small
    \begin{tabular}{>{\centering\arraybackslash}p{1.4cm}}
        \hline
        Method  \\ 
        \hline 
        CLIP  \\
        GPT-4V \\
        DINOv2 \\
        ResNet-50 \\
        SSCD  \\
        BoT  \\
        \hline 
        Label \\
        \hline 
    \end{tabular}
\end{minipage}
\begin{minipage}{0.115\linewidth}
    \centering
    \includegraphics[height=3cm]{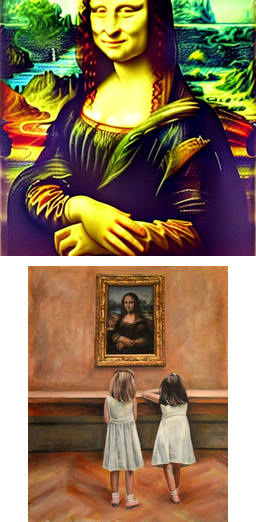}
\end{minipage}
\hspace{0.1cm}
\begin{minipage}{0.07\linewidth}
    \centering
    \small
    \begin{tabular}{>{\centering\arraybackslash}p{0.6cm}}
        \hline
        Sim. \\ 
        \hline 
        0.63 \\
        0.40 \\
        0.35 \\
        0.54 \\
        0.04 \\
        0.05 \\
        \hline 
        1.00 \\
        \hline 
    \end{tabular}
\end{minipage}
\hspace{0.1cm}
\begin{minipage}{0.115\linewidth}
    \centering
    \includegraphics[height=3cm]{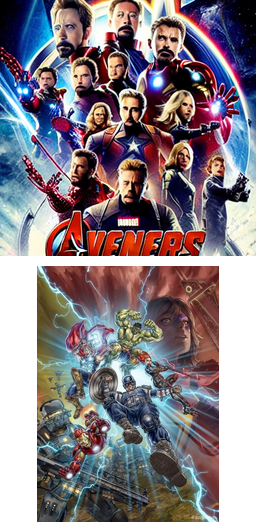}
\end{minipage}
\hspace{0.1cm}
\begin{minipage}{0.07\linewidth}
    \centering
    \small
    \begin{tabular}{>{\centering\arraybackslash}p{0.6cm}}
        \hline
        Sim. \\ 
        \hline 
        0.68 \\
        0.60 \\
        0.65 \\
        0.61 \\
        0.16 \\
        0.36 \\
        \hline 
        1.00 \\
        \hline 
    \end{tabular}
\end{minipage}
\hspace{0.1cm}
\begin{minipage}{0.115\linewidth}
    \centering
    \includegraphics[height=3cm]{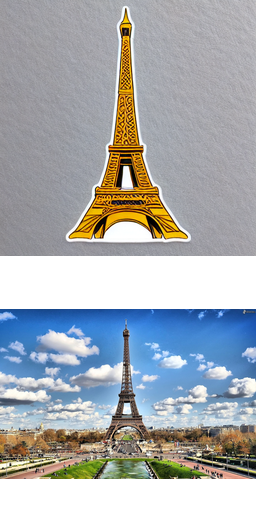}
\end{minipage}
\hspace{0.1cm}
\begin{minipage}{0.07\linewidth}
    \centering
    \small
    \begin{tabular}{>{\centering\arraybackslash}p{0.6cm}}
        \hline
        Sim. \\ 
        \hline 
        0.85 \\
        0.30 \\
        0.69 \\
        0.53 \\
        0.08 \\
        0.21 \\
        \hline 
        1.00 \\
        \hline 
    \end{tabular}
\end{minipage}
\hspace{0.1cm}
\begin{minipage}{0.115\linewidth}
    \centering
    \includegraphics[height=3cm]{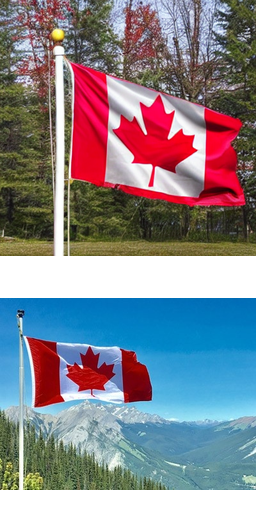}
\end{minipage}
\hspace{0.1cm}
\begin{minipage}{0.07\linewidth}
    \centering
    \small
    \begin{tabular}{>{\centering\arraybackslash}p{0.6cm}}
        \hline
        Sim. \\ 
        \hline 
        0.87 \\
        0.90 \\
        0.35 \\
        0.58 \\
        0.10 \\
        0.06 \\
        \hline 
        1.00 \\
        \hline 
    \end{tabular}
\end{minipage}

    \vspace{0.21cm}
    
    \begin{minipage}{0.13\linewidth}
    \centering
    \small
    \begin{tabular}{>{\centering\arraybackslash}p{1.4cm}}
        \hline
        Method  \\ 
        \hline 
        CLIP  \\
        GPT-4V \\
        DINOv2 \\
        ResNet-50 \\
        SSCD  \\
        BoT  \\
        \hline 
        Label \\
        \hline 
    \end{tabular}
\end{minipage}
\begin{minipage}{0.115\linewidth}
    \centering
    \includegraphics[height=3cm]{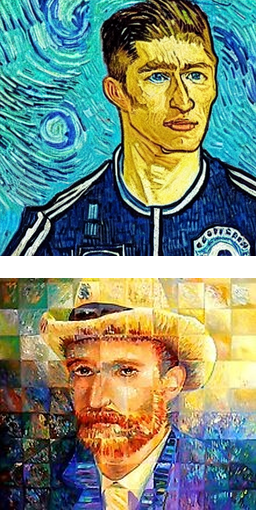}
\end{minipage}
\hspace{0.1cm}
\begin{minipage}{0.07\linewidth}
    \centering
    \small
    \begin{tabular}{>{\centering\arraybackslash}p{0.6cm}}
        \hline
        Sim. \\ 
        \hline 
        0.49 \\
        0.70 \\
        0.40 \\
        0.67 \\
        0.11 \\
        0.24 \\
        \hline 
        0.80 \\
        \hline 
    \end{tabular}
\end{minipage}
\hspace{0.1cm}
\begin{minipage}{0.115\linewidth}
    \centering
    \includegraphics[height=3cm]{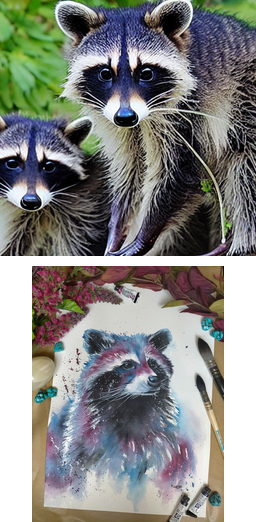}
\end{minipage}
\hspace{0.1cm}
\begin{minipage}{0.07\linewidth}
    \centering
    \small
    \begin{tabular}{>{\centering\arraybackslash}p{0.6cm}}
        \hline
        Sim. \\ 
        \hline 
        0.57 \\
        0.20 \\
        0.60 \\
        0.56 \\
        0.13 \\
        0.36 \\
        \hline 
        0.80 \\
        \hline 
    \end{tabular}
\end{minipage}
\hspace{0.1cm}
\begin{minipage}{0.115\linewidth}
    \centering
    \includegraphics[height=3cm]{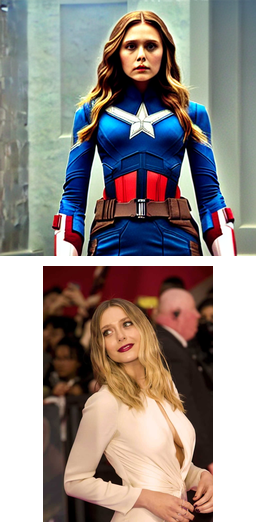}
\end{minipage}
\hspace{0.1cm}
\begin{minipage}{0.07\linewidth}
    \centering
    \small
    \begin{tabular}{>{\centering\arraybackslash}p{0.6cm}}
        \hline
        Sim. \\ 
        \hline 
        0.65 \\
        0.00 \\
        0.21 \\
        0.65 \\
        0.00 \\
        0.11 \\
        \hline 
        0.80 \\
        \hline 
    \end{tabular}
\end{minipage}
\hspace{0.1cm}
\begin{minipage}{0.115\linewidth}
    \centering
    \includegraphics[height=3cm]{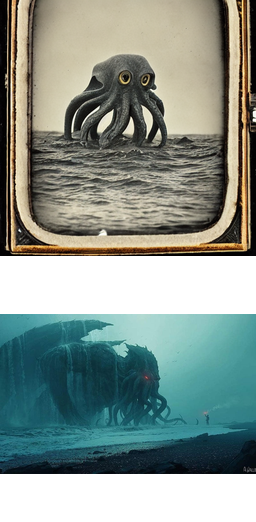}
\end{minipage}
\hspace{0.1cm}
\begin{minipage}{0.07\linewidth}
    \centering
    \small
    \begin{tabular}{>{\centering\arraybackslash}p{0.6cm}}
        \hline
        Sim. \\ 
        \hline 
        0.68 \\
        0.30 \\
        0.44 \\
        0.56 \\
        0.12 \\
        0.25 \\
        \hline 
        0.80 \\
        \hline 
    \end{tabular}
\end{minipage}

    \vspace{0.21cm}
    
    \begin{minipage}{0.13\linewidth}
    \centering
    \small
    \begin{tabular}{>{\centering\arraybackslash}p{1.4cm}}
        \hline
        Method  \\ 
        \hline 
        CLIP  \\
        GPT-4V \\
        DINOv2 \\
        ResNet-50 \\
        SSCD  \\
        BoT  \\
        \hline 
        Label \\
        \hline 
    \end{tabular}
\end{minipage}
\begin{minipage}{0.115\linewidth}
    \centering
    \includegraphics[height=3cm]{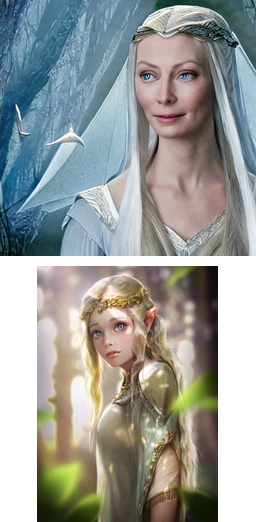}
\end{minipage}
\hspace{0.1cm}
\begin{minipage}{0.07\linewidth}
    \centering
    \small
    \begin{tabular}{>{\centering\arraybackslash}p{0.6cm}}
        \hline
        Sim. \\ 
        \hline 
        0.70 \\
        0.70 \\
        0.61 \\
        0.60 \\
        0.08 \\
        0.25 \\
        \hline 
        0.60 \\
        \hline 
    \end{tabular}
\end{minipage}
\hspace{0.1cm}
\begin{minipage}{0.115\linewidth}
    \centering
    \includegraphics[height=3cm]{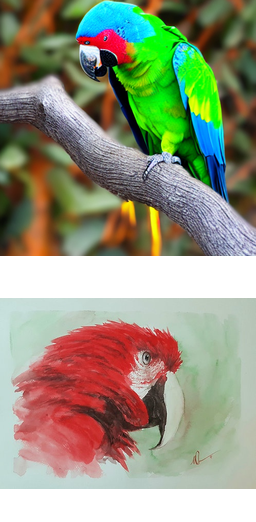}
\end{minipage}
\hspace{0.1cm}
\begin{minipage}{0.07\linewidth}
    \centering
    \small
    \begin{tabular}{>{\centering\arraybackslash}p{0.6cm}}
        \hline
        Sim. \\ 
        \hline 
        0.71 \\
        0.20 \\
        0.59 \\
        0.60 \\
        0.02 \\
        0.03 \\
        \hline 
        0.60 \\
        \hline 
    \end{tabular}
\end{minipage}
\hspace{0.1cm}
\begin{minipage}{0.115\linewidth}
    \centering
    \includegraphics[height=3cm]{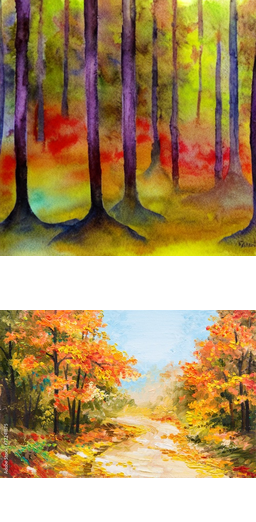}
\end{minipage}
\hspace{0.1cm}
\begin{minipage}{0.07\linewidth}
    \centering
    \small
    \begin{tabular}{>{\centering\arraybackslash}p{0.6cm}}
        \hline
        Sim. \\ 
        \hline 
        0.82 \\
        0.30 \\
        0.40 \\
        0.57 \\
        0.08 \\
        0.22 \\
        \hline 
        0.60 \\
        \hline 
    \end{tabular}
\end{minipage}
\hspace{0.1cm}
\begin{minipage}{0.115\linewidth}
    \centering
    \includegraphics[height=3cm]{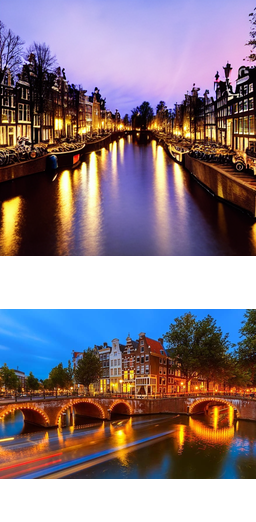}
\end{minipage}
\hspace{0.1cm}
\begin{minipage}{0.07\linewidth}
    \centering
    \small
    \begin{tabular}{>{\centering\arraybackslash}p{0.6cm}}
        \hline
        Sim. \\ 
        \hline 
        0.93 \\
        0.80 \\
        0.76 \\
        0.62 \\
        0.31 \\
        0.35 \\
        \hline 
        0.60 \\
        \hline 
    \end{tabular}
\end{minipage}

    \vspace{0.21cm}
    
    \begin{minipage}{0.13\linewidth}
    \centering
    \small
    \begin{tabular}{>{\centering\arraybackslash}p{1.4cm}}
        \hline
        Method  \\ 
        \hline 
        CLIP  \\
        GPT-4V \\
        DINOv2 \\
        ResNet-50 \\
        SSCD  \\
        BoT  \\
        \hline 
        Label \\
        \hline 
    \end{tabular}
\end{minipage}
\begin{minipage}{0.115\linewidth}
    \centering
    \includegraphics[height=3cm]{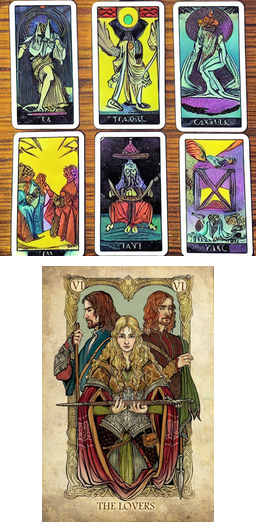}
\end{minipage}
\hspace{0.1cm}
\begin{minipage}{0.07\linewidth}
    \centering
    \small
    \begin{tabular}{>{\centering\arraybackslash}p{0.6cm}}
        \hline
        Sim. \\ 
        \hline 
        0.57 \\
        0.20 \\
        0.28 \\
        0.67 \\
        0.01 \\
        0.21 \\
        \hline 
        0.40 \\
        \hline 
    \end{tabular}
\end{minipage}
\hspace{0.1cm}
\begin{minipage}{0.115\linewidth}
    \centering
    \includegraphics[height=3cm]{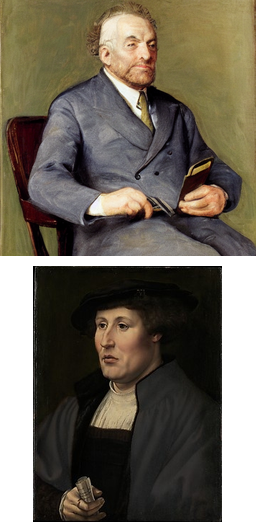}
\end{minipage}
\hspace{0.1cm}
\begin{minipage}{0.07\linewidth}
    \centering
    \small
    \begin{tabular}{>{\centering\arraybackslash}p{0.6cm}}
        \hline
        Sim. \\ 
        \hline 
        0.57 \\
        0.30 \\
        0.58 \\
        0.60 \\
        0.14 \\
        0.21 \\
        \hline 
        0.40 \\
        \hline 
    \end{tabular}
\end{minipage}
\hspace{0.1cm}
\begin{minipage}{0.115\linewidth}
    \centering
    \includegraphics[height=3cm]{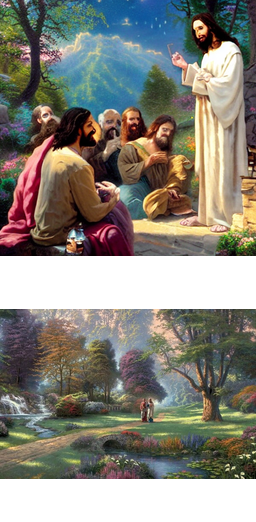}
\end{minipage}
\hspace{0.1cm}
\begin{minipage}{0.07\linewidth}
    \centering
    \small
    \begin{tabular}{>{\centering\arraybackslash}p{0.6cm}}
        \hline
        Sim. \\ 
        \hline 
        0.62 \\
        0.10 \\
        0.23 \\
        0.63 \\
        0.15 \\
        0.25 \\
        \hline 
        0.40 \\
        \hline 
    \end{tabular}
\end{minipage}
\hspace{0.1cm}
\begin{minipage}{0.115\linewidth}
    \centering
    \includegraphics[height=3cm]{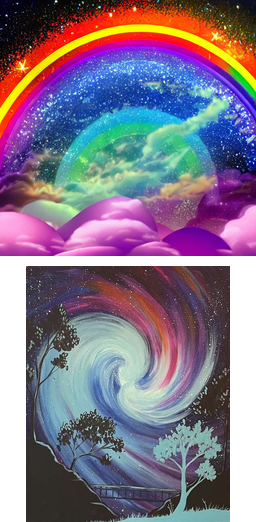}
\end{minipage}
\hspace{0.1cm}
\begin{minipage}{0.07\linewidth}
    \centering
    \small
    \begin{tabular}{>{\centering\arraybackslash}p{0.6cm}}
        \hline
        Sim. \\ 
        \hline 
        0.77 \\
        0.30 \\
        0.55 \\
        0.66 \\
        0.10 \\
        0.16 \\
        \hline 
        0.40 \\
        \hline 
    \end{tabular}
\end{minipage}

\vspace{0.21cm}

\begin{minipage}{0.13\linewidth}
    \centering
    \small
    \begin{tabular}{>{\centering\arraybackslash}p{1.4cm}}
        \hline
        Method  \\ 
        \hline 
        CLIP  \\
        GPT-4V \\
        DINOv2 \\
        ResNet-50 \\
        SSCD  \\
        BoT  \\
        \hline 
        Label \\
        \hline 
    \end{tabular}
\end{minipage}
\begin{minipage}{0.115\linewidth}
    \centering
    \includegraphics[height=3cm]{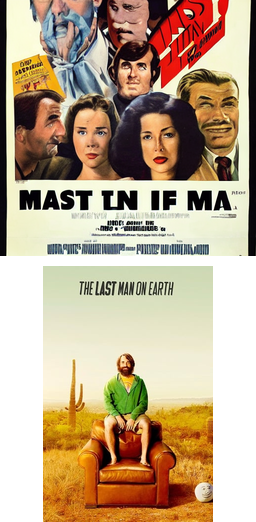}
\end{minipage}
\hspace{0.1cm}
\begin{minipage}{0.07\linewidth}
    \centering
    \small
    \begin{tabular}{>{\centering\arraybackslash}p{0.6cm}}
        \hline
        Sim. \\ 
        \hline 
        0.39 \\
        0.00 \\  
        0.07 \\
        0.59 \\
        0.06 \\
        0.15 \\
        \hline 
        0.20 \\
        \hline 
    \end{tabular}
\end{minipage}
\hspace{0.1cm}
\begin{minipage}{0.115\linewidth}
    \centering
    \includegraphics[height=3cm]{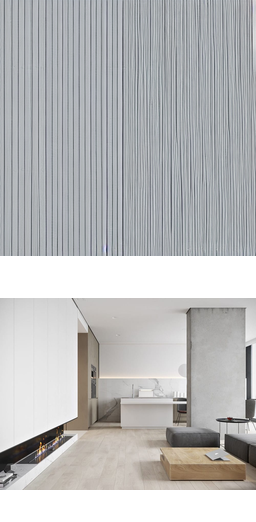}
\end{minipage}
\hspace{0.1cm}
\begin{minipage}{0.07\linewidth}
    \centering
    \small
    \begin{tabular}{>{\centering\arraybackslash}p{0.6cm}}
        \hline
        Sim. \\ 
        \hline 
        0.55 \\
        0.00 \\
        0.17 \\
        0.59 \\
        0.02 \\
        0.01 \\
        \hline 
        0.20 \\
        \hline 
    \end{tabular}
\end{minipage}
\hspace{0.1cm}
\begin{minipage}{0.115\linewidth}
    \centering
    \includegraphics[height=3cm]{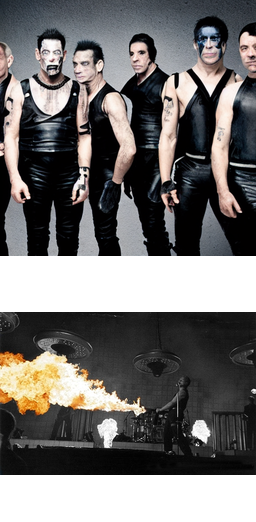}
\end{minipage}
\hspace{0.1cm}
\begin{minipage}{0.07\linewidth}
    \centering
    \small
    \begin{tabular}{>{\centering\arraybackslash}p{0.6cm}}
        \hline
        Sim. \\ 
        \hline 
        0.56 \\
        0.00 \\
        0.11 \\
        0.59 \\
        0.05 \\
        0.05 \\
        \hline 
        0.20 \\
        \hline 
    \end{tabular}
\end{minipage}
\hspace{0.1cm}
\begin{minipage}{0.115\linewidth}
    \centering
    \includegraphics[height=3cm]{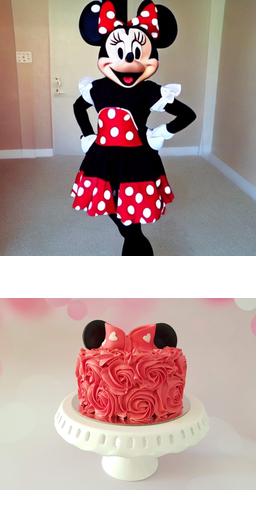}
\end{minipage}
\hspace{0.1cm}
\begin{minipage}{0.07\linewidth}
    \centering
    \small
    \begin{tabular}{>{\centering\arraybackslash}p{0.6cm}}
        \hline
        Sim. \\ 
        \hline 
        0.69 \\
        0.00\\
        0.11 \\
        0.56 \\
        0.12 \\
        0.05 \\
        \hline 
        0.20 \\
        \hline 
    \end{tabular}
\end{minipage}

\vspace{0.21cm}

\begin{minipage}{0.13\linewidth}
    \centering
    \small
    \begin{tabular}{>{\centering\arraybackslash}p{1.4cm}}
        \hline
        Method  \\ 
        \hline 
        CLIP  \\
        GPT-4V \\
        DINOv2 \\
        ResNet-50 \\
        SSCD  \\
        BoT  \\
        \hline 
        Label \\
        \hline 
    \end{tabular}
\end{minipage}
\begin{minipage}{0.115\linewidth}
    \centering
    \includegraphics[height=3cm]{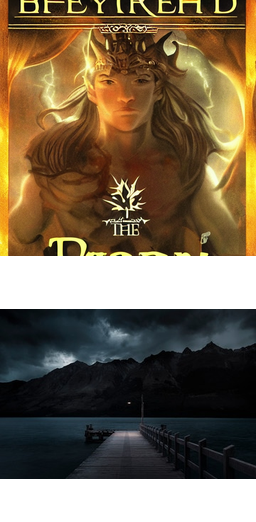}
\end{minipage}
\hspace{0.1cm}
\begin{minipage}{0.07\linewidth}
    \centering
    \small
    \begin{tabular}{>{\centering\arraybackslash}p{0.6cm}}
        \hline
        Sim. \\ 
        \hline 
        0.38 \\
        0.00 \\
        0.06 \\
        0.61 \\
        0.05 \\
        0.12 \\
        \hline 
        0.00 \\
        \hline 
    \end{tabular}
\end{minipage}
\hspace{0.1cm}
\begin{minipage}{0.115\linewidth}
    \centering
    \includegraphics[height=3cm]{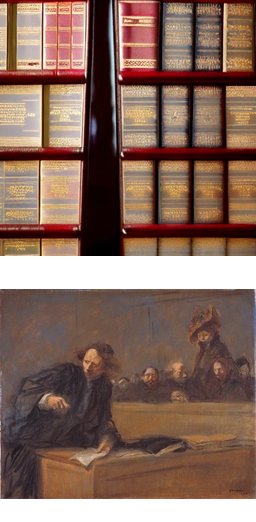}
\end{minipage}
\hspace{0.1cm}
\begin{minipage}{0.07\linewidth}
    \centering
    \small
    \begin{tabular}{>{\centering\arraybackslash}p{0.6cm}}
        \hline
        Sim. \\ 
        \hline 
        0.56 \\
        0.00 \\
        0.02 \\
        0.62 \\
        0.05 \\
        0.08 \\
        \hline 
        0.00 \\
        \hline 
    \end{tabular}
\end{minipage}
\hspace{0.1cm}
\begin{minipage}{0.115\linewidth}
    \centering
    \includegraphics[height=3cm]{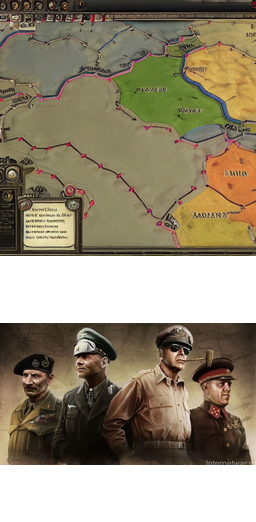}
\end{minipage}
\hspace{0.1cm}
\begin{minipage}{0.07\linewidth}
    \centering
    \small
    \begin{tabular}{>{\centering\arraybackslash}p{0.6cm}}
        \hline
        Sim. \\ 
        \hline 
        0.57 \\
        0.00 \\
        0.09 \\
        0.53 \\
        0.04 \\
        0.03 \\
        \hline 
        0.00 \\
        \hline 
    \end{tabular}
\end{minipage}
\hspace{0.1cm}
\begin{minipage}{0.115\linewidth}
    \centering
    \includegraphics[height=3cm]{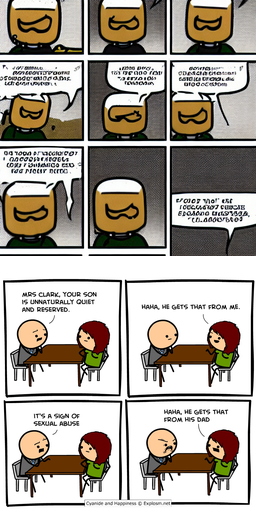}
\end{minipage}
\hspace{0.1cm}
\begin{minipage}{0.07\linewidth}
    \centering
    \small
    \begin{tabular}{>{\centering\arraybackslash}p{0.6cm}}
        \hline
        Sim. \\ 
        \hline 
        0.78 \\
        0.10 \\
        0.38 \\
        0.63 \\
        0.23 \\
        0.43 \\
        \hline 
        0.00 \\
        \hline 
    \end{tabular}
\end{minipage}
 \vspace{3mm}
\captionof{figure}{The similarities (or normalized levels) predicted by existing models.} 

    \label{Fig: sim}
\end{figure}

\begin{figure}[H]
\centering
    \vspace{6mm}
    \includegraphics[width=14cm]{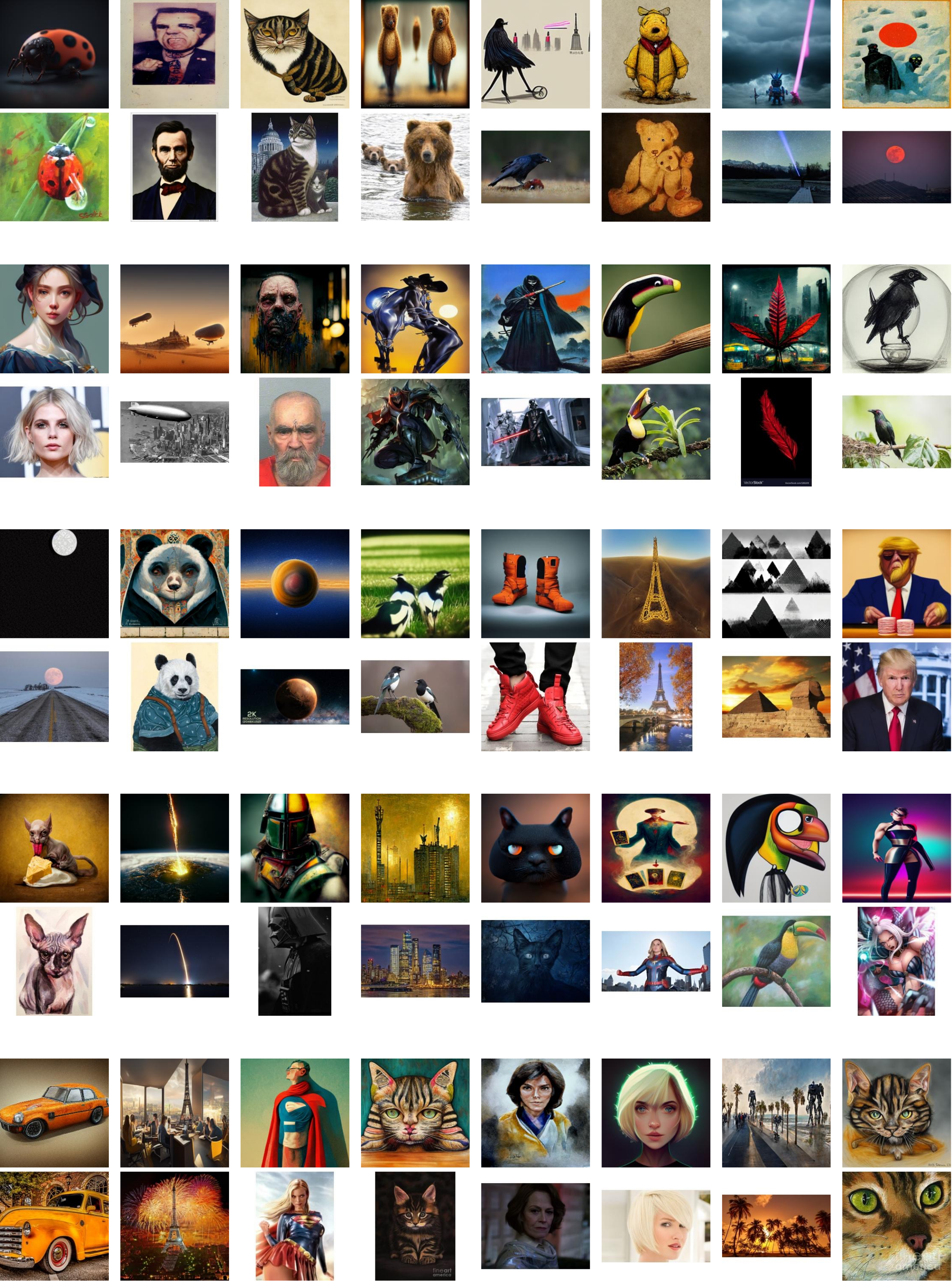}
    \vspace{3mm}
    \captionof{figure}{The replication examples generated by Midjourney \cite{midjourney2022}.} 

    \label{Fig: midall}
\end{figure}

\begin{figure}[H]
\centering
\vspace{6mm}
    \includegraphics[width=14cm]{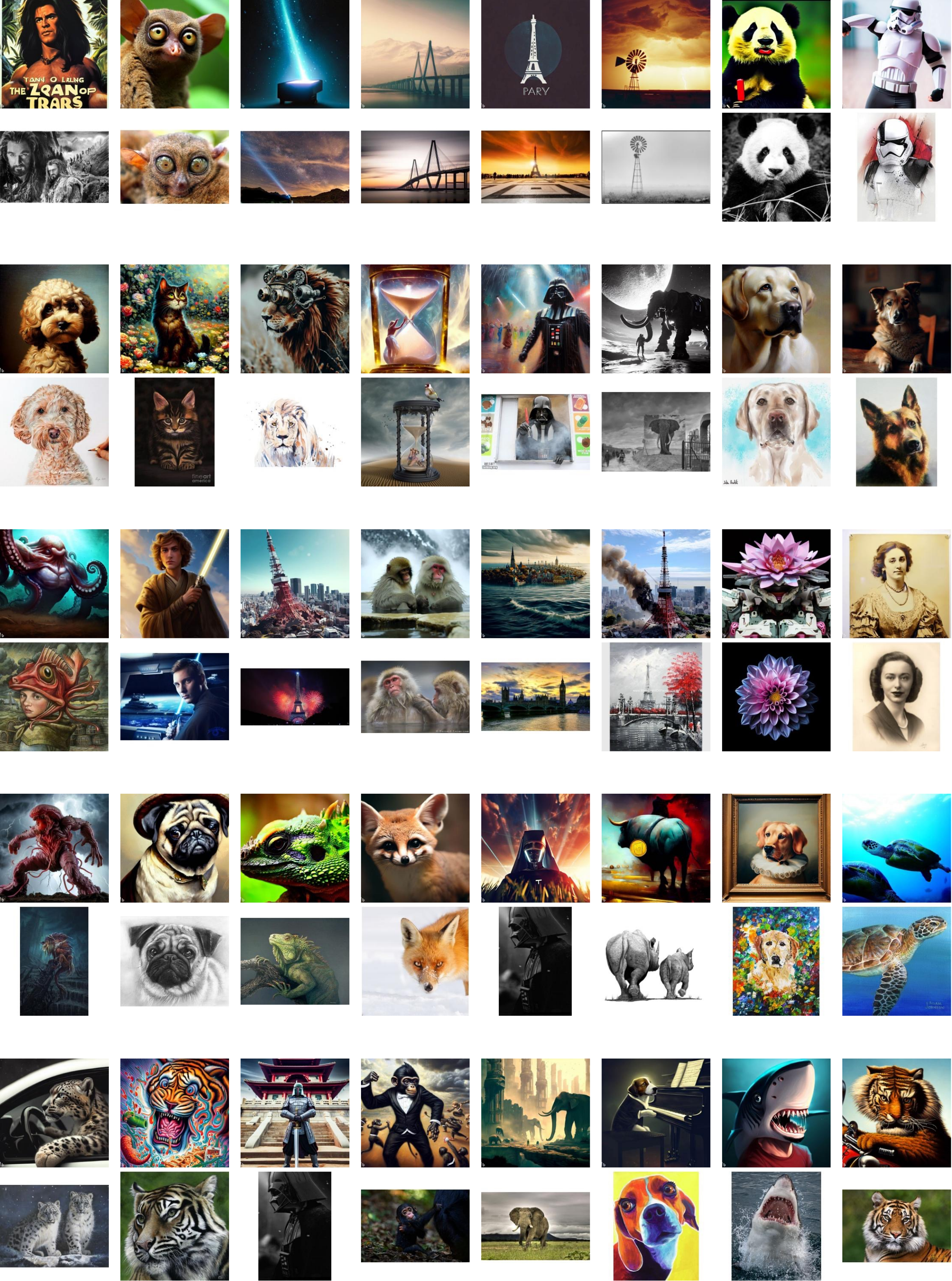}
    \vspace{3mm}
    \captionof{figure}{The replication examples generated by New Bing \cite{new_bing}.} 

    \label{Fig: bingall}
\end{figure}

\begin{figure}[H]
\centering
\vspace{6mm}
    \includegraphics[width=14cm]{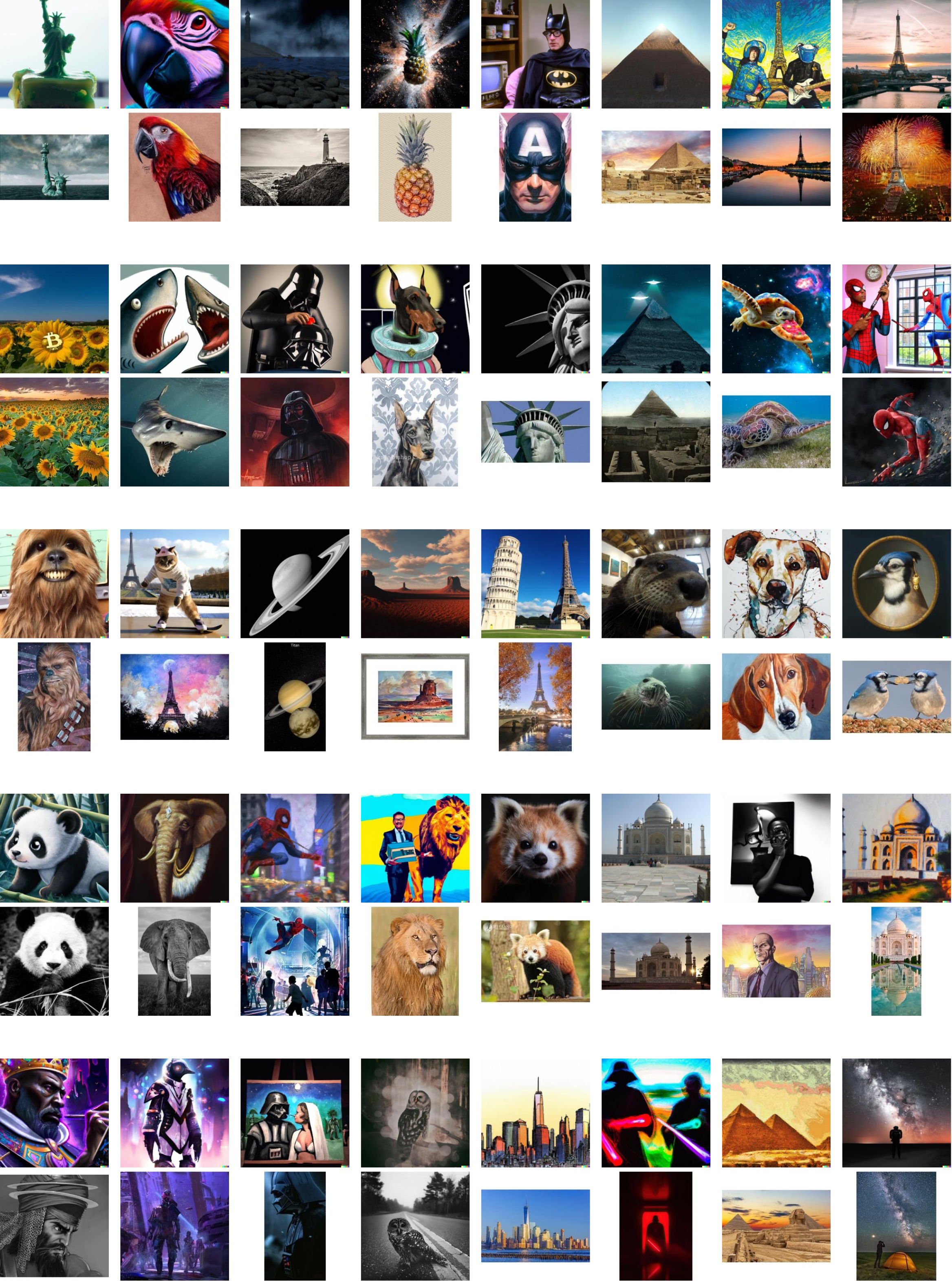}
    \vspace{3mm}
    \captionof{figure}{The replication examples generated by DALLE·2 \cite{ramesh2022hierarchical}.} 

    \label{Fig: dalleall}
\end{figure}

\begin{figure}[H]
\centering
\vspace{6mm}
    \includegraphics[width=14cm]{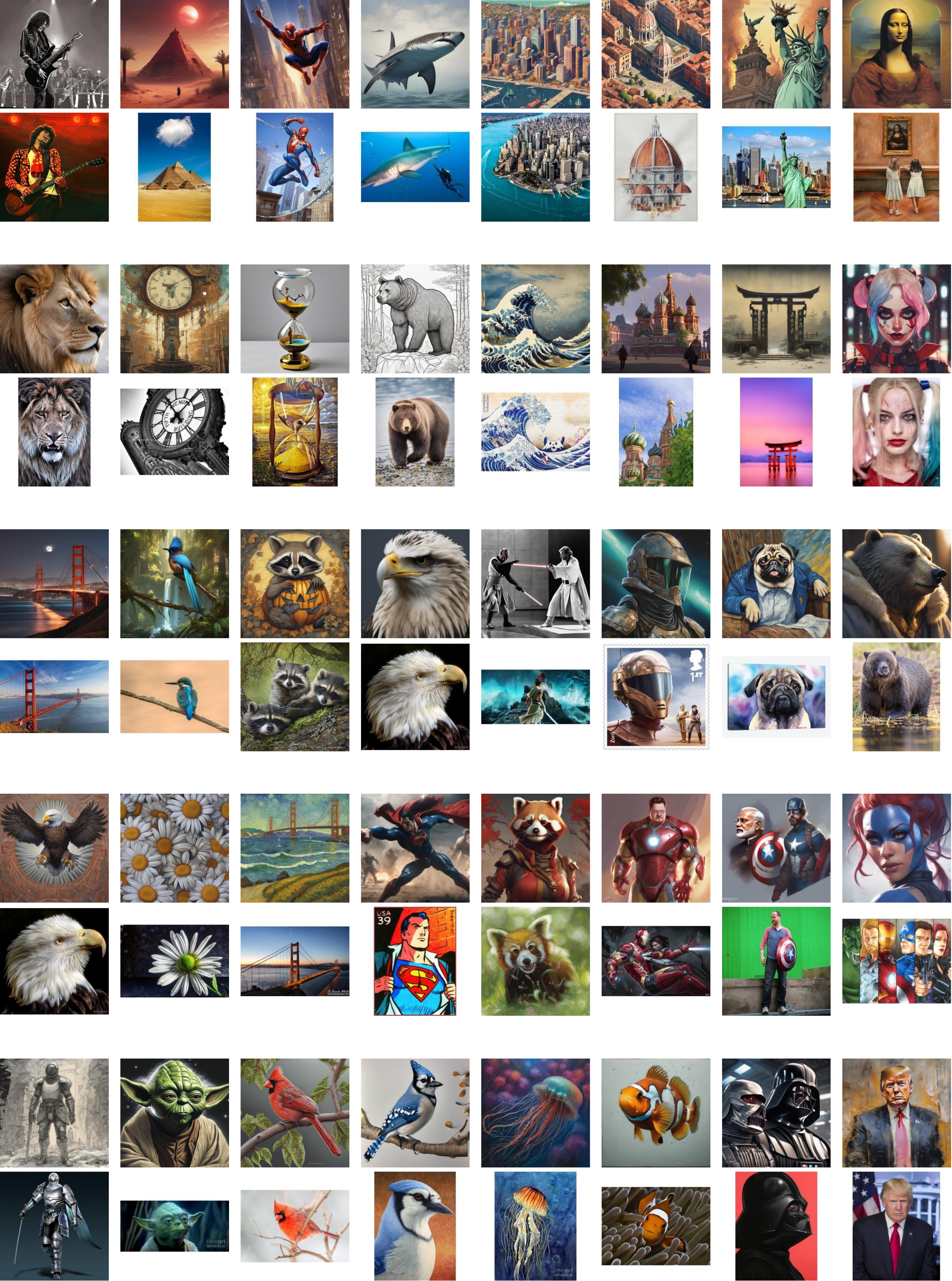}
    \vspace{3mm}
    \captionof{figure}{The replication examples generated by SDXL \cite{podell2023sdxl}.} 

    \label{Fig: sdxlall}
\end{figure}

\begin{figure}[H]
\centering
\vspace{6mm}
    \includegraphics[width=14cm]{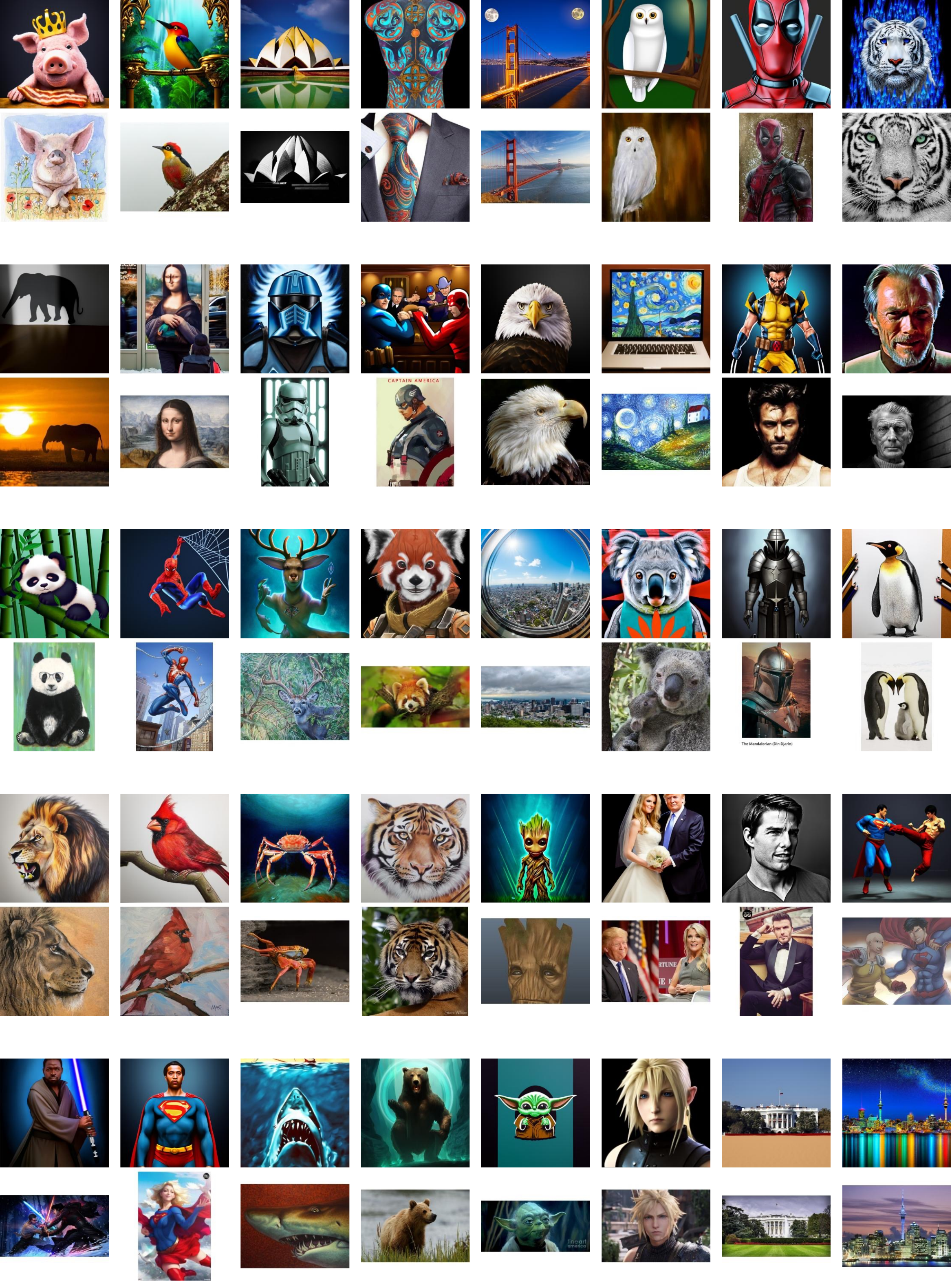}
    \vspace{3mm}
    \captionof{figure}{The replication examples generated by DeepFloyd IF \cite{deep_floyd_IF}.} 

    \label{Fig: floydeall}
\end{figure}

\begin{figure}[H]
\centering
\vspace{6mm}
    \includegraphics[width=14cm]{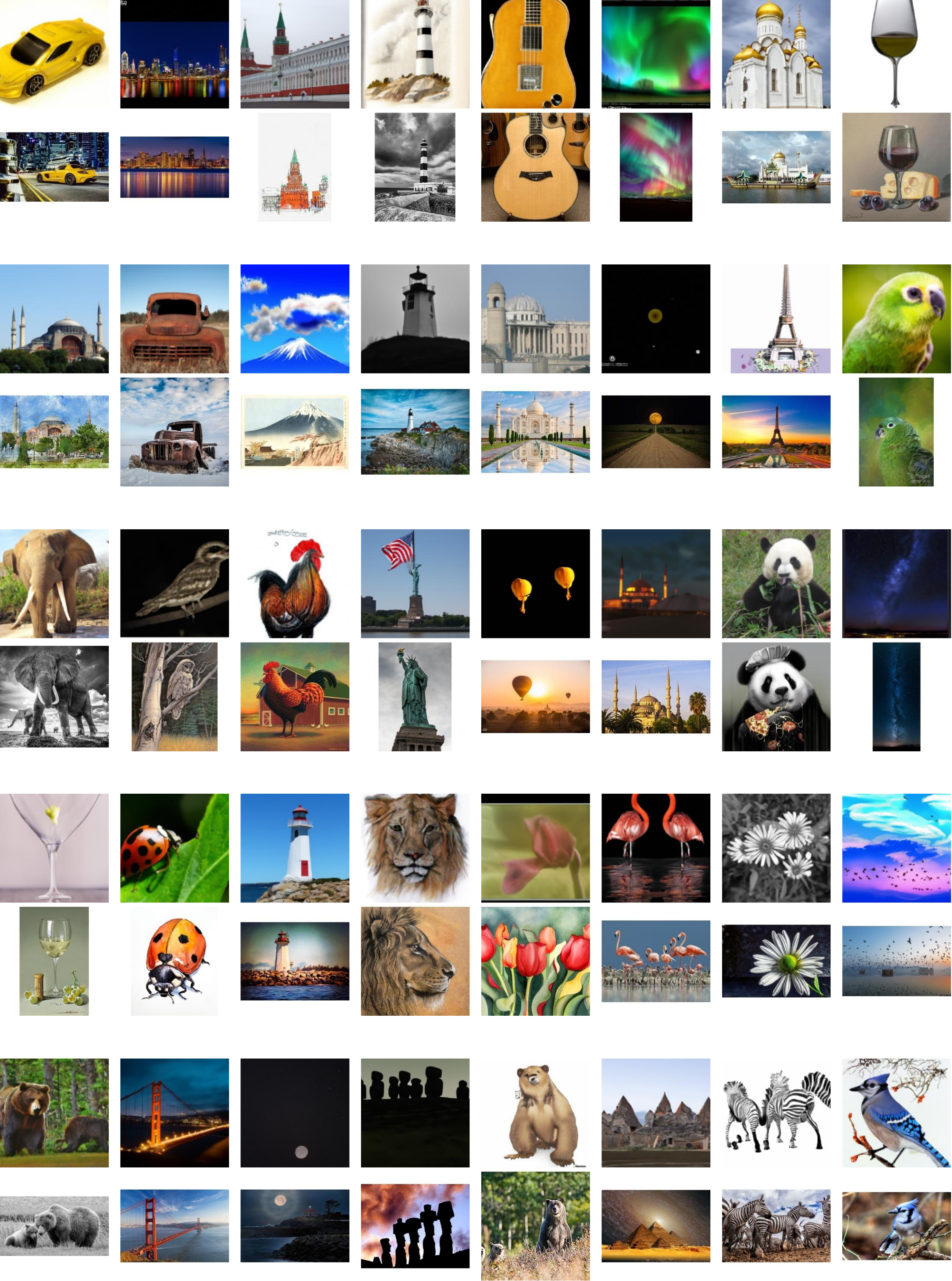}
    \vspace{3mm}
    \captionof{figure}{The replication examples generated by GLIDE \cite{nichol2022glide}.} 

    \label{Fig: glideall}
\end{figure}

\end{document}